\pdfoutput=1
\documentclass[runningheads]{llncs}
\usepackage{graphicx}
\usepackage{amsmath,amssymb} 
\usepackage{color}
\usepackage{multirow}
\usepackage{array}
\usepackage{arydshln}
\usepackage{makecell}

\begin{document}
\pagestyle{headings}
\mainmatter

\def\ACCV20SubNumber{630}  

\title{Synthetic-to-real domain adaptation for lane detection}  
\titlerunning{Synthetic-to-real domain adaptation for lane detection}
%
\author{Noa Garnett \and Roy Uziel \and Netalee Efrat \and Dan Levi}
\authorrunning{Garnett et al.}
%
\institute{General Motors Technical Center Israel - R\&D Labs}

\maketitle

\begin{abstract}
Accurate lane detection, a crucial enabler for autonomous driving,  currently relies on obtaining a large and diverse labeled training dataset. In this work, we explore learning from abundant, randomly generated synthetic data, together with unlabeled or partially labeled target domain data, instead. Randomly generated synthetic data has the advantage of controlled variability in the lane geometry and lighting, but it is limited in terms of photo-realism. This poses the challenge of adapting models learned on the unrealistic synthetic domain to real images. To this end we develop a novel autoencoder-based approach that uses synthetic labels unaligned with particular images for adapting to target domain data. In addition, we explore existing domain adaptation approaches, such as image translation and self-supervision, and adjust them to the lane detection task. We test all approaches in the unsupervised domain adaptation setting in which no target domain labels are available and in the semi-supervised setting in which a small portion of the target images are labeled. In extensive experiments using three different datasets, we demonstrate the possibility to save costly target domain labeling efforts. For example, using our proposed autoencoder approach on the llamas and tuSimple lane datasets, we can almost recover the fully supervised accuracy with only 10\% of the labeled data. In addition, our autoencoder approach outperforms all other methods in the semi-supervised domain adaptation scenario.
\end{abstract}
\section{Introduction}
Accurate lane detection is a critical enabler for autonomous driving, serving as a primary input for the path planning stage. All current state-of-the-art implementations involve training a single-frame CNN based detector~\cite{Hou:ICCV:2019:learning,Garnett:cvpr:2019:3d,tuSimple}. The real-world success of the resulting detector relies on the assumption that the training set reliably represents the operational conditions. This poses the challenge of collecting and labeling images that represent all possible driving scenarios: from highway to urban scenes, in all weather and lighting conditions, covering all lane marking types from all the different geographic regions. Labeling such a large and diverse corpus of data is a highly demanding task to say the least. While several highly-diverse on-road datasets were collected and made public, many of them lack lane annotations (e.g. Waymo open dataset~\cite{waymo_dataset}, nuScenes~\cite{Cesar:arxiv:2019:nuscenes}). Synthetic datasets have also been proposed for training autonomous perception models~\cite{Gaidon:cvpr:2016:virtual},\cite{vKITTI}. Unfortunately, the scene creation is still a manual labor demanding task, but more importantly, lanes are often not inserted as graphical objects but as road texture, and therefore no lane annotation can be automatically extracted.

\begin{figure}
    \centering
    \begin{tabular}{c|c}
    \textbf{(a)} & \textbf{(b)} \\
    \begin{tabular}{cc}
         \includegraphics[height=2cm]{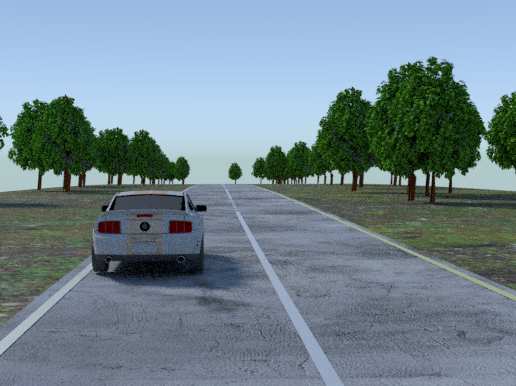}& \includegraphics[height=2cm]{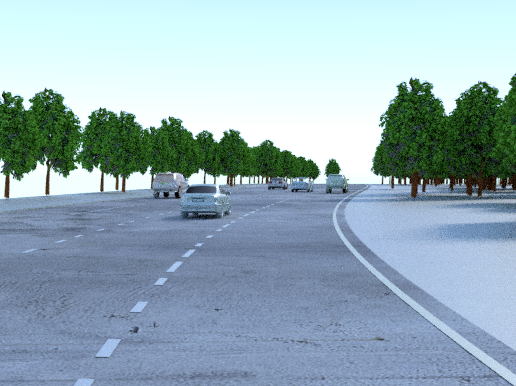}  \\
         \includegraphics[height=2cm]{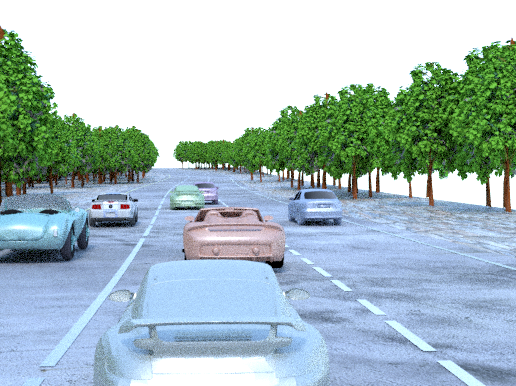}& \includegraphics[height=2cm]{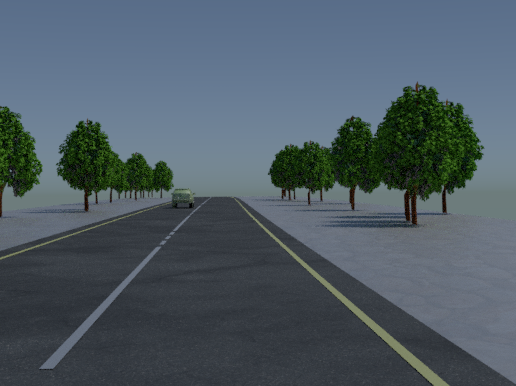} \\
          \includegraphics[height=2cm]{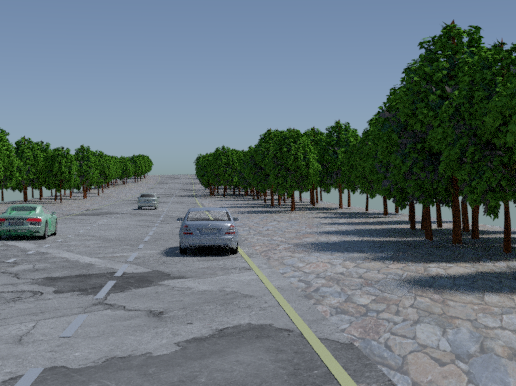}& \includegraphics[height=2cm]{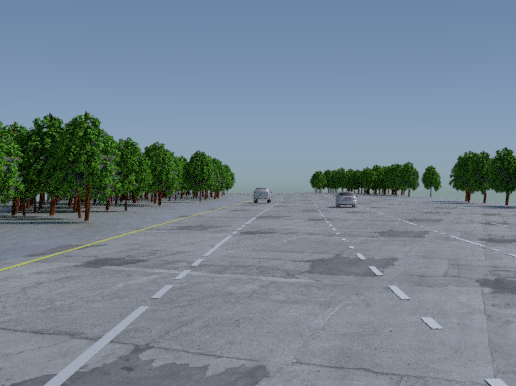} \\
    \end{tabular}         &
    \begin{tabular}{c}
    \includegraphics[height=3.2cm]{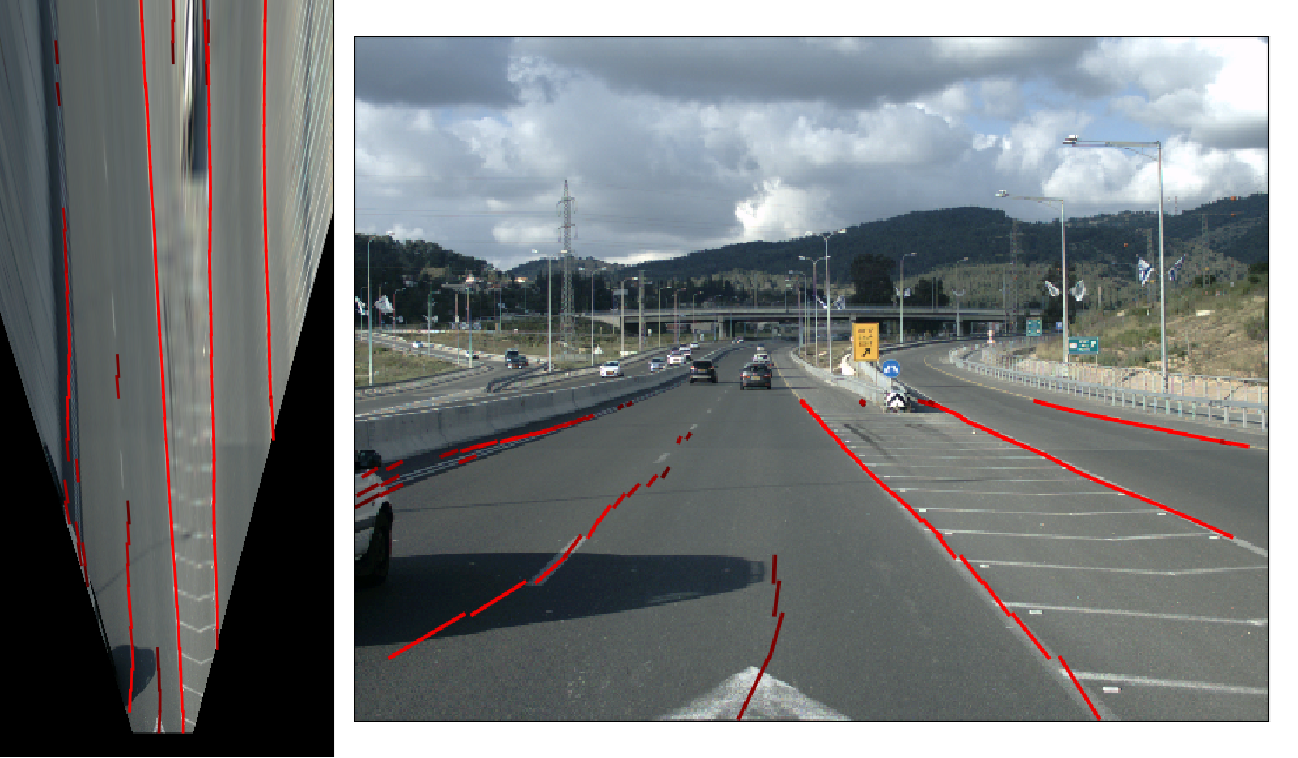} \\
    \hline
     \includegraphics[height=3.2cm]{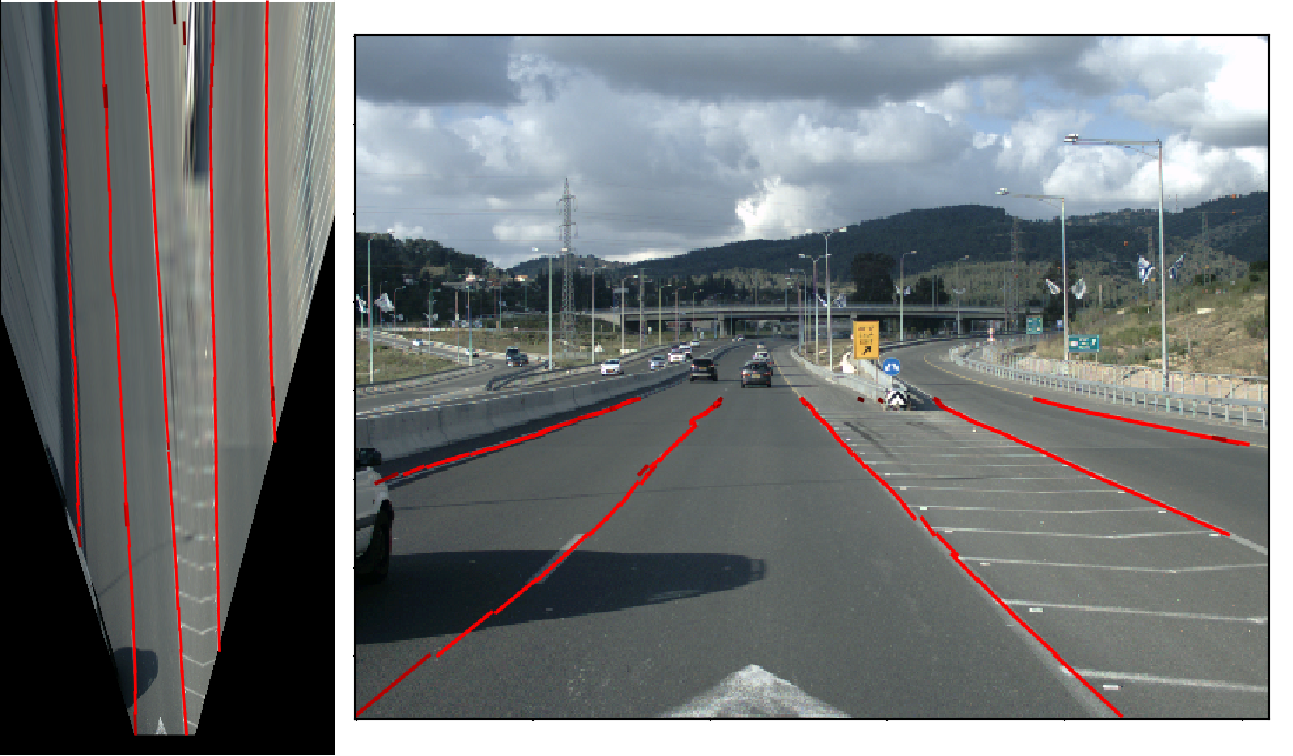}
    \end{tabular}
    \end{tabular}
    \caption{\textbf{(a)} Examples of images from our source domain, consisting of synthetic images randomly generated using the methodology of~\cite{Garnett:cvpr:2019:3d}.\textbf{ (b)} Lane detection results on an example from the 3DLanes dataset~\cite{netalee}. \textbf{Top:} result of a model trained with synthetic data without domain adaptation. \textbf{Bottom:} After domain adaptation using the self-supervision objective described in Section~\ref{sec:method}. Each result is shown in top-view, as obtained directly from our network, and back-projected to regular view on the right. The result is shown by highlighting in red the detected lane segments for tiles with confidence above a minimum threshold. The brightness of the segment color reflects its confidence score.}
    \label{fig:intro}
\end{figure}

Recently,~\cite{Garnett:cvpr:2019:3d} introduced a methodology for randomly and efficiently generating synthetic data with lane annotation. This approach has the advantage of generating scenes with high variability in lane topology and 3D geometry, without requiring any manual labor. The caveat lies in the limited variability in the appearance of the lanes and the surrounding scene, and generally in the photo-realism of the resulting images, as can be seen in Figure~\ref{fig:intro}(a). We propose to leverage this synthetic data generation method, along with unlabeled real-world data, and to perform synthetic-to-real domain adaptation (DA) in order to overcome the appearance gap. This would allow exploiting the variability in lane geometry provided by the synthetic data along with the appearance variability of unlabeled real data.

Although domain adaptation is a well-studied field, it has yet to be applied to lane detection. In this work, we introduce a novel autoencoder-based approach and a new self-supervision objective addressing domain adaptation for the lane detection task. In addition, we explore existing domain adaptation methods by adjusting them to lane detection. We address domain adaptation in two different settings: unsupervised and semi-supervised. In the Unsupervised Domain Adaptation (UDA) setting, training is done with only source domain labeled data and target domain unlabeled data.  In the Semi-Supervised Domain Adaptation (SSDA) setting training has additionally access to a small set of labeled examples from the target domain. In both settings, the goal is to learn an accurate model for the target domain, with minimal compromise compared to a fully supervised model.

Our proposed autoencoder-based method, is inspired by a recent study~\cite{jakab2019learning}, showing that it is feasible to learn human landmark detection from only unlabeled images and unpaired labels. Similarly, we introduce a method that learns to detect lanes using unlabeled images, and a set of ground truth ''logical lane images'' (Figure~\ref{fig:arch}(b)), which are not paired with input images. This allows to train a lane detector without even having rendered synthetic data, but with only ``logical'' top view images of valid lane markings. Based on the assumption that there is a strong correlation between the lanes in a scene and the gradients in the image, we train an autoencoder of the image gradients that passes through a ``lane image'' bottleneck. Using adversarial training, we force the appearance of the resulting lane image to resemble that of the ground truth lane images. 

Self-supervision leverages automatically generated supervision on auxiliary tasks related to the objective task to force the creation of useful intermediate network representations. To this end, we introduce a new self-supervision task, which improves performance on lane detection.  We then use target domain self-supervision together with source domain supervised training for domain adaptation. Finally, we implemented three additional approaches that require some adjustment to the lane detection task: image-to-image translation~\cite{Hoffman:arxiv:2017:Cycada}, feature distribution alignment using adversarial learning~\cite{Ganin_ICML15} and central moment discrepancy~\cite{Zellinger:arxiv:2017:cmd}.
We evaluate all methods on three datasets: tuSimple~\cite{tuSimple}, llamas~\cite{Nehrendt:ICCV:2019:llamas} and 3DLanes~\cite{netalee}. We show that applying domain adaptation, using several of the tested methods, significantly improves performance on the target domain, as can be observed in Figure~\ref{fig:intro}(b). We also show that it improves performance in the semi-supervised setting. In particular, our autoencoder approach nearly closes the accuracy gap on two out of the three datasets compared to full supervision with only a tenth of the labels. To summarize, our main contributions are:
\begin{itemize}
\item Showing that synthetic-to-real domain adaptation can significantly improve lane detection performance when there is little or no target domain labels
\item Introducing a novel autoencoder-based DA approach for lane detection that achieves state-of-the-art performance in the SSDA setting on all tested datasets
\item Introducing a new self-supervision objective for lane detection
\item Providing a comprehensive comparison and  evaluation of proposed and existing methods in both UDA and SSDA settings.
\end{itemize}
\section{Related work}


There has been extensive prior work on unsupervised domain adaptation. The general idea is to align the distributions of the source and target domains at some level of the representation. This can be accomplished by two general strategies. The first strategy attempts to align the two domains at an intermediate feature-level representation. One way to achieve this is to directly minimize the discrepancy between the two distributions using measures such as maximum mean discrepancy (MMD)~\cite{Bousmalis_NIPS2016,Long_ICML15} and central moment distance (CMD)~\cite{Zellinger:arxiv:2017:cmd}. Another approach for bringing representations closer is based on adversarial learning~\cite{Ganin_ICML15,Adda_CVPR2017}. The idea is to train an adversarial discriminator that distinguishes between the two domains, encouraging the generator, in this case, the feature embedding network, to generate indistinguishable feature maps. Feature embedding weights may be shared between the domains~\cite{Ganin_ICML15} or kept separated~\cite{Adda_CVPR2017}. Several additional variants of this approach have been proposed: ~\cite{Long_NIPS2018_7436} condition the adversarial game not only on the embedding but also on the final output,~\cite{lahiri2018unsupervised} align the features at multiple levels, and ~\cite{Gen2Adapt} use the discriminator over images generated from the embedding. Our implementation for this approach (\textbf{Embedding GAN}) uses a adversarial learning on the feature embedding and shared weights as in~\cite{Ganin_ICML15}.

The second strategy uses image-to-image translation to impose alignment at the raw image level. Image translation methods \cite{Zhu:iccv:2017:cycle,Huang:iccv:2017:adain} learn the mapping between two image domains, and is mainly used for style transfer, object transfiguration, season transfer, and photo enhancement. In the context of domain adaptation, the image translation generates a target domain image from a raw source domain one, allowing to train a model in the supervised setting on a "target-like" dataset~\cite{TaigmanPW17,Hoffman:arxiv:2017:Cycada,BousmalisSDEK17,CrDoCo}. As shown in \cite{Hoffman:arxiv:2017:Cycada}, the two strategies can be combined to improve performance further. For our task we implement \textbf{image translation} using CycleGAN~\cite{Zhu:iccv:2017:cycle}.  

Self-supervised learning uses an auxiliary task generated automatically from the data itself to train feature representations that would hopefully be useful for the end-goal task. Many auxiliary tasks have been proposed, such as jigsaw puzzle solving~\cite{NorooziF16}, image rotation prediction~\cite{GidarisSK18}, and contrastive predictive coding~\cite{oord2018representation}. Typically, self-supervision is applied to large sets of unlabeled data as a pre-train stage before supervised training~\cite{hnaff2019dataefficient}. Naturally, as concurrently proposed by~\cite{sun2019unsupervised}, self-supervision can be used for unsupervised domain adaptation. The idea is to train the feature embedding on the supervised task using the labeled source dataset, and on the auxiliary task using the unlabeled target images. Assuming the correlation between the tasks, the hope is that the feature embedding will encode the correlated information similarly for both domains. In ~\cite{sun2019unsupervised}, the feature embedding is further aligned by training the self-supervision task on the source domain.

Compared to the research in domain adaptation for \textit{classification},  much less attention has been paid to domain adaptation for other, more complex, computer vision tasks. Several studies address object detection~\cite{chen2018domain,zhu2019adapting}, monocular depth estimation~\cite{AbarghoueiB18,zheng2018t2net} and semantic segmentation~\cite{Hoffman:arxiv:2017:Cycada,Hong_CVPR18,Zhang_2018,Sankaranarayanan18,CrDoCo}. To our best knowledge, domain adaptation for lane detection has yet to be addressed. While synthetic-to-real adaptation for automotive perception has been previously studied~\cite{Chen_CVPR19,Hoffman:arxiv:2017:Cycada,Hong_CVPR18,Zhang_2018,Sankaranarayanan18}, it was always with manually created synthetic datasets (e.g. vKITTI~\cite{vKITTI}, SYNTHIA~\cite{synthia}) as opposed to the random generation approach we adopt from~\cite{Garnett:cvpr:2019:3d}.

\section{Methods}
\label{sec:method}

\begin{figure}
\centering
\begin{tabular}{c:c}
    (a) & (b) \\
     \includegraphics[height=5.5cm]{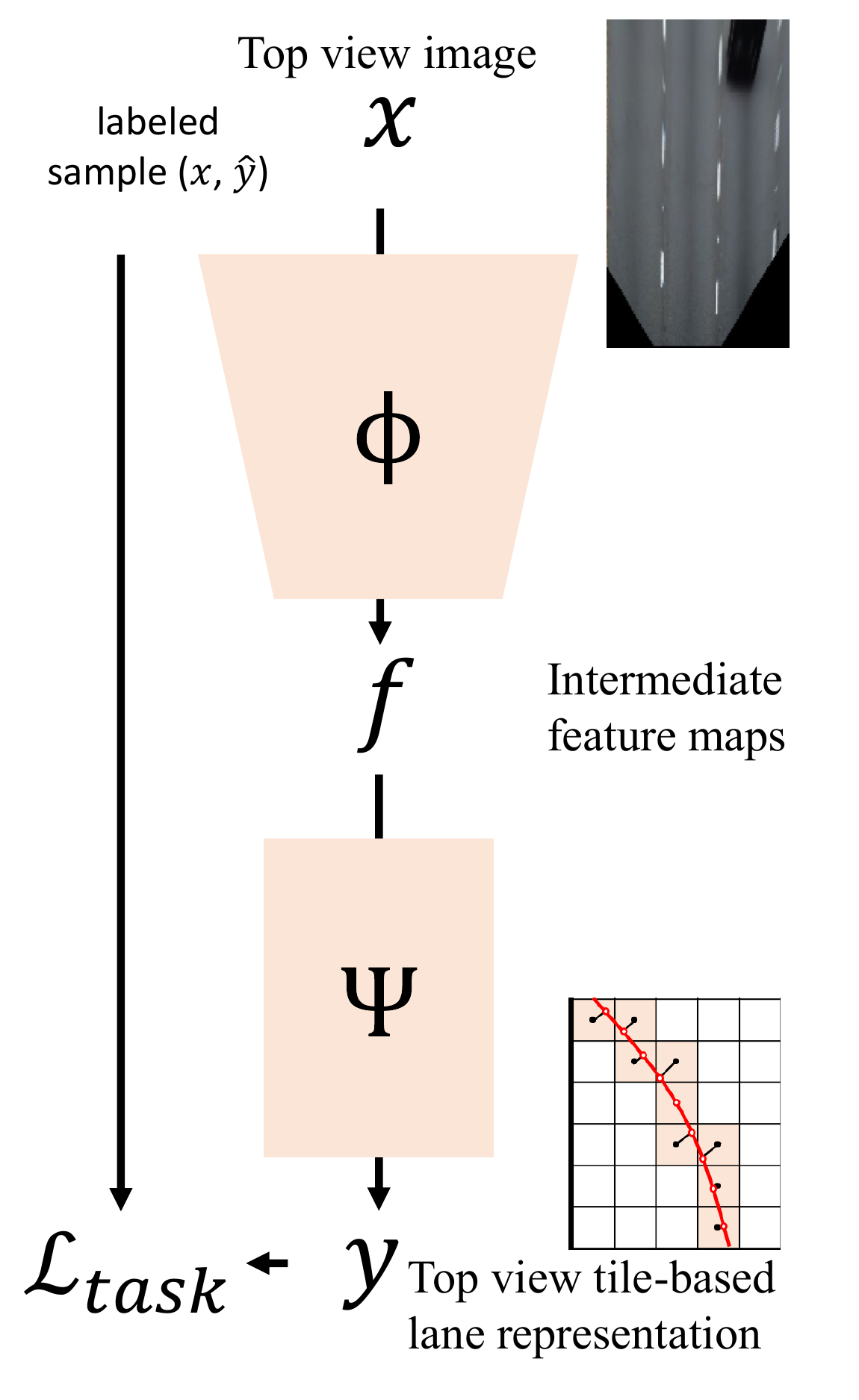} & 
     \includegraphics[height=5.5cm]{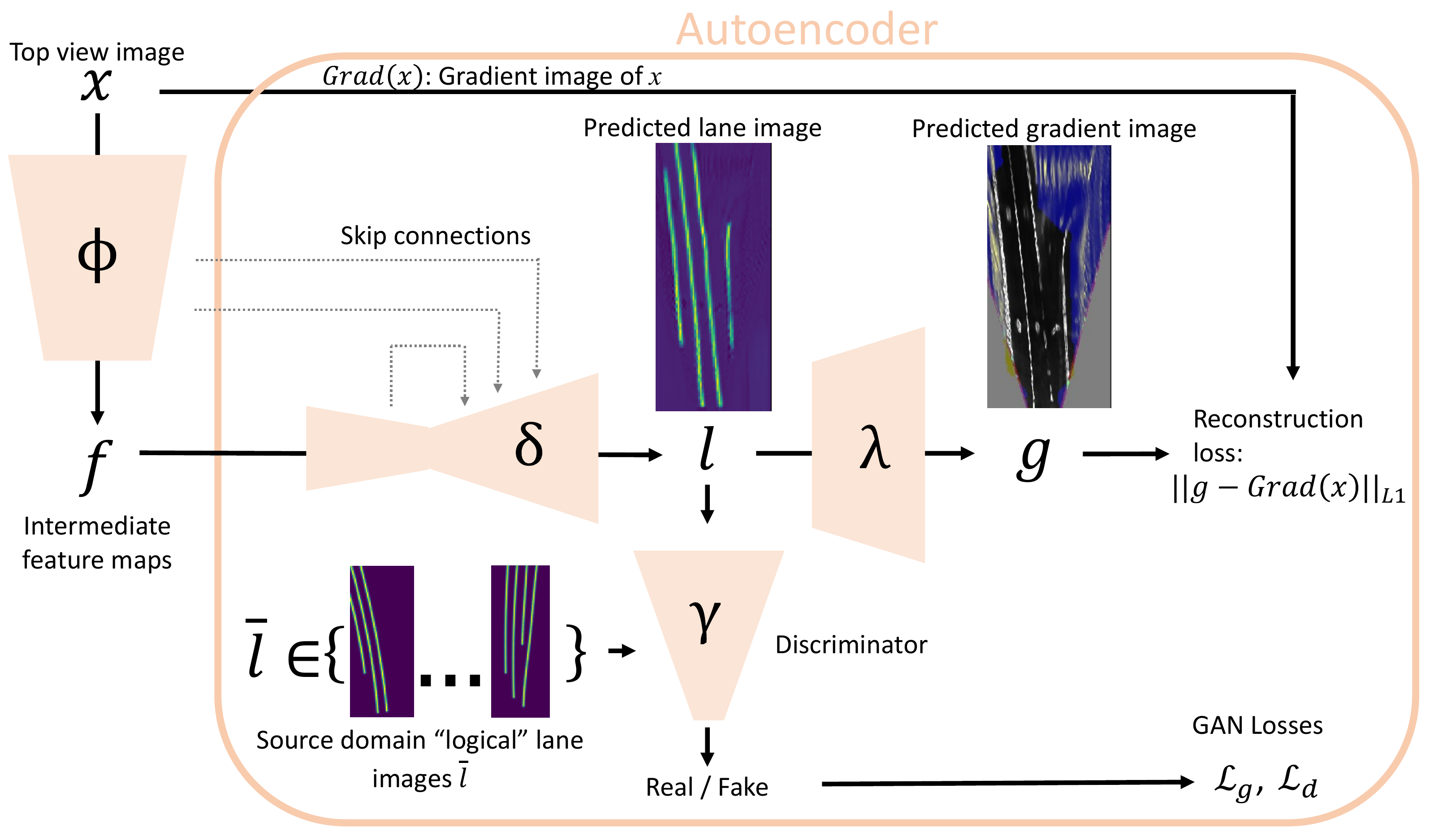}
\end{tabular}

\caption{\textbf{(a) Base architecture for lane detection.} $f=\phi(x)$  are intermediate feature maps computed from the top view. $y=\psi(f)$ represents the detected lanes in the tiles representation~\cite{netalee}. This base architecture, used also in inference, is trained end to end in the fully supervised setting. \textbf{(b) Our proposed autoencoder architecture for training with unlabeled examples and unpaired top view lane images.} The networks $\phi,\delta,\lambda$ are trained to minimize the reconstruction loss while $\delta\circ\phi$ and $\gamma$ are adverserially trained as a generator and discriminator respectively, driving the generated lane image $l$ to resemble ground truth annotations $\overline{l}$.  Note that $\delta\circ\phi$ forms an hourglass architecture including skip connections. In $g$ the blue pixels are not considered in the reconstruction loss using the heuristics described in Section $\ref{sec:method}$. At inference the network can output $\delta\circ\phi(x)$, which may be additionally transformed to the tile representation using a pretrained transformer network (See Section~\ref{sec:experiments} for more details).}
\label{fig:arch}
\end{figure}

We start by warping the original image $I$ to a virtual top view image $x\in\mathcal{X}=\mathbb{R}^{3\times H' \times W'}$ using an Inverse Perspective Mapping (IPM), which is a homography defined by the camera's position relative to the local ground plane, its intrinsic parameters and additional parameters embodied by $H'$ and $W'$. Our lane detection network gets $x$ and outputs a lane representation $y$ describing the lanes in top view. Working in top view has the advantage of translation invariance, an important property for convolutional networks, and is also beneficial to subsequent modules such as lane clustering and tracking~\cite{Garnett:cvpr:2019:3d,netalee}. We use the semi-local tile-based representation, recently proposed in~\cite{netalee}, that divides the top view into $H\times W$ non-overlapping tiles, each roughly corresponding $1.6$ by $1.6$ meters in the real world. For each tile $(i,j)$ the network outputs the confidence ($b_{(i,j)}\in [ 0, 1 ]$) that there is a lane passing through the tile, and regresses using a set of continuous outputs $\mathbf{p}_{(i,j)}\in \mathbb{R}^9$ its position and orientation relative to the rectangle's center, assuming that locally the lane is linear. We follow the exact representation of~\cite{netalee} except for omitting the elevation offset which is used for detection in 3D. The output is then expressed as: $y=\{(b_{(i,j)},\mathbf{p}_{(i,j)})|i\in [1,\ldots,H], j\in[1,\ldots, W]\}\in \mathcal{Y}=R^{10\times H \times W}$. 

Our base network architecture, used in inference, is illustrated in Figure~\ref{fig:arch}(a). It is convenient to present the base network as composed of two stages. First, an embedding convolutional neural network $\phi:\mathcal{X} \mapsto \mathcal {F}=\mathbb{R}^{C\times H \times W}$ generates an intermediate feature map representation $f$, with the same spatial dimensions as the output, and $C$ channels. Then, an additional convolutional network $\psi:\mathcal{F} \mapsto \mathcal{Y} $ computes the output $y$ from the intermediate feature maps $f$. Given a dataset  $\mathcal{D}^l$ consisting of labeled examples $(x,\hat{y})\in{\mathcal{X,Y}}$, the supervised task function loss is:
\begin{equation*}
    \mathcal{L}_{task}=\sum_{(x,
    \hat{y})\in \mathcal{D}^l}\mathcal{L}_{tiles}(\psi\circ\phi(x),\hat{y})
\end{equation*} 
Where $\mathcal{L}_{tiles}(y,\hat{y})$ sums the loss across all tiles for a single output $y$ and ground truth $\hat{y}$ as described in~\cite{netalee} omitting the elevation component. Note also that $\hat{y}$ is obtained by projecting \textit{image annotations} to top-view using the same IPM applied to the image.

In the \textbf{fully supervised} setting, we simply have a labeled dataset $\mathcal{D}^l$ and train the base network with the task loss function $\mathcal{L}_{task}$. In the \textbf{unsupervised domain adaptation} (UDA) setting, we have two datasets, a labeled source domain dataset $\mathcal{D}^l_S$, and an unlabeled set of images $\mathcal{D}^u_T$ from the target domain. Finally, in the \textbf{semi-supervised domain adaptation} (SSDA) setting we have in addition, compared to the UDA setting, a small set of labeled target domain images $\mathcal{D}^l_T$. We next describe our proposed autoencoder (Section~\ref{sec:AE}) and self-supervision (Section~\ref{sec:Self}) approaches followed by a short description of the existing methods we additionally tested in Sections~\ref{sec:IT} and \ref{sec:DM}.

\begin{figure}[ht]
\centering
\begin{tabular}{ccc}
    (a) \textbf{Autoencoder} & (b) \textbf{Self-supervision} \\
     \includegraphics[height=4.3cm]{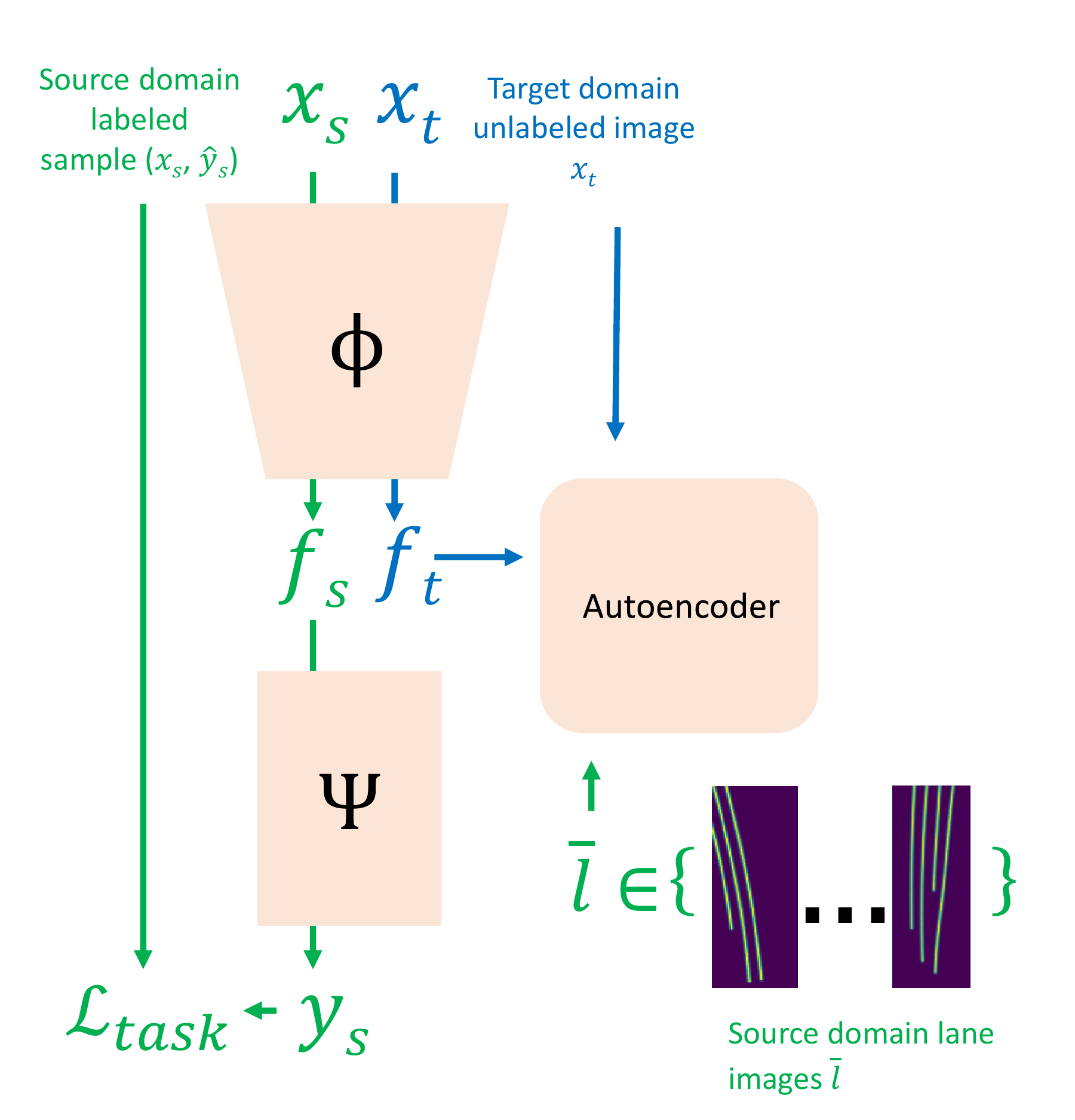} & 
     \includegraphics[height=4.3cm]{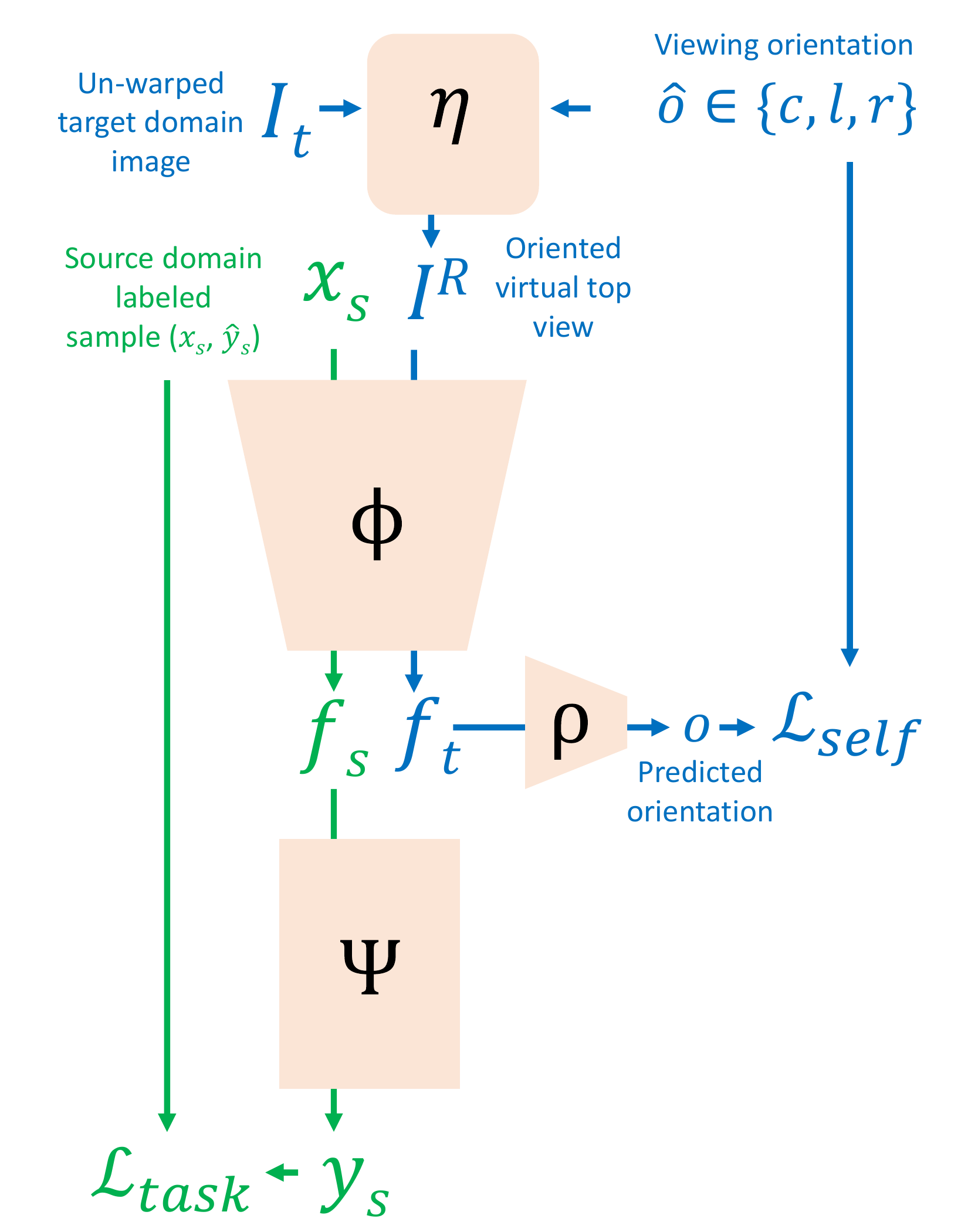} \\
    (c) \textbf{Image translation} & (d) \textbf{Embedding GAN} \\
     \includegraphics[height=4.3cm]{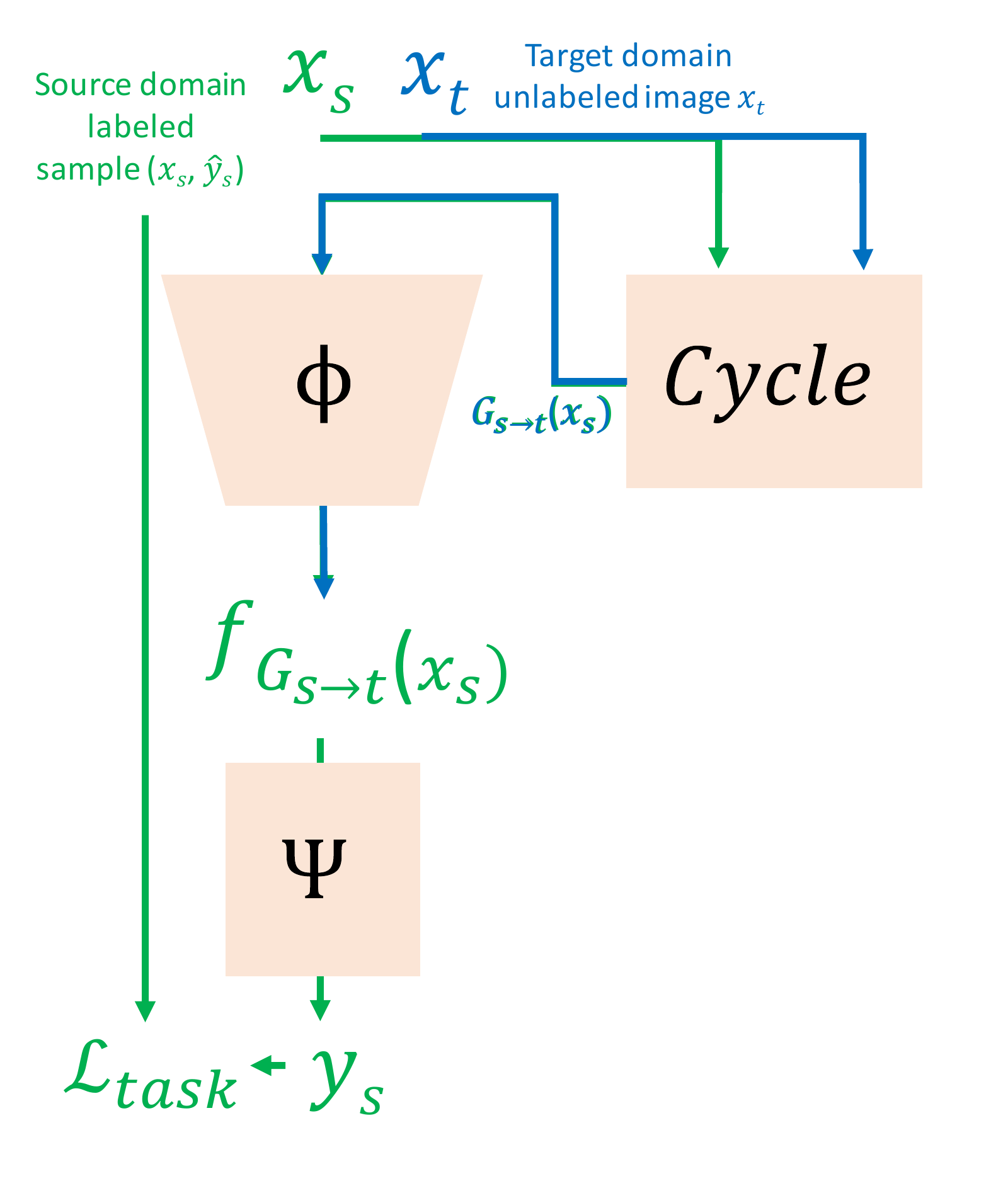} & 
     \includegraphics[height=4.3cm]{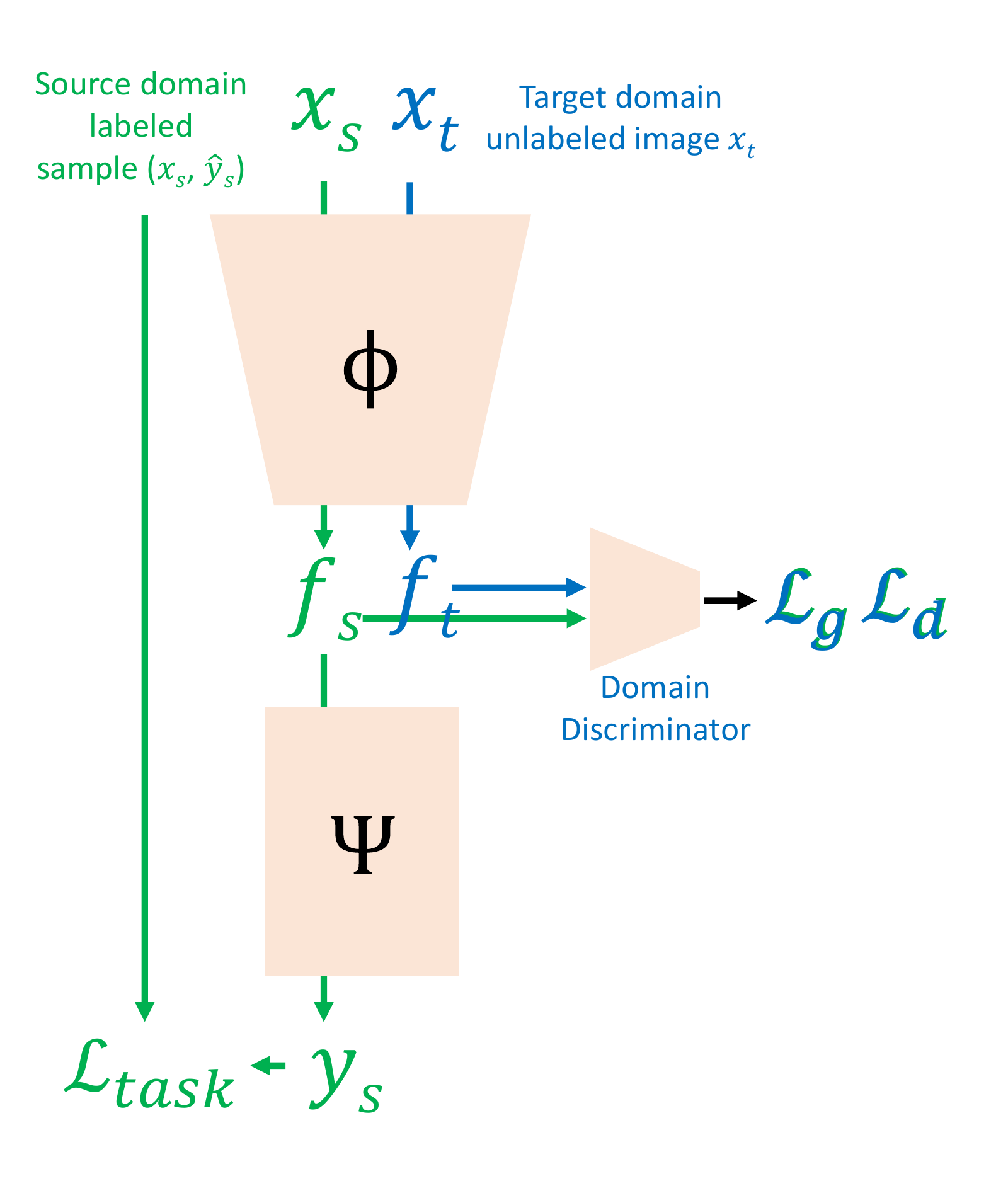}
\end{tabular}

\caption{\textbf{Architectures for the different domain adaptation approaches in the UDA setting}. See Section~\ref{sec:method} for further details.}
\end{figure}~\label{fig:da_arch}
\begin{figure}[ht]
\centering
\begin{tabular}{ccc}
    left & center & right \\
     \includegraphics[height=4cm]{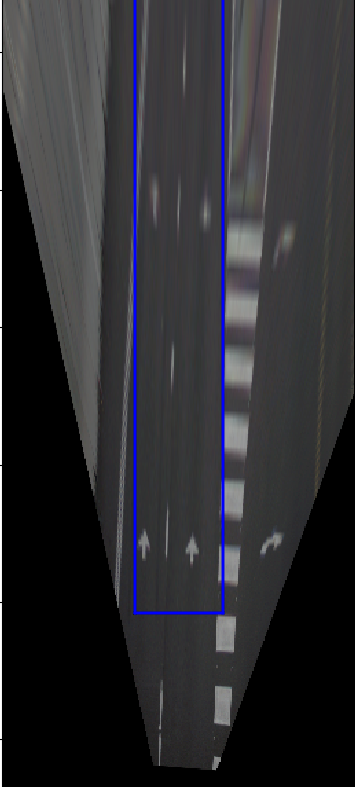} &

     \includegraphics[height=4cm]{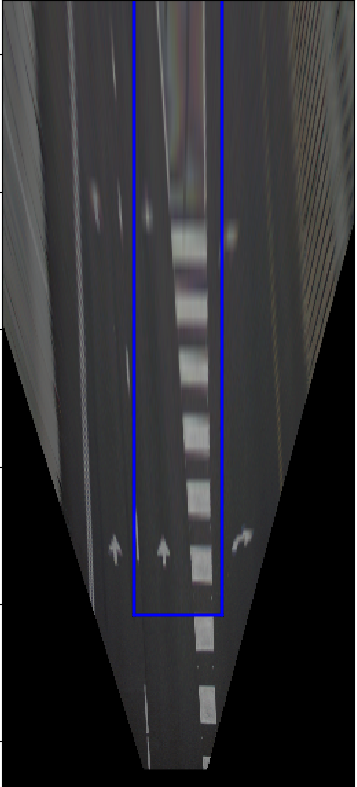} &
     \includegraphics[height=4cm]{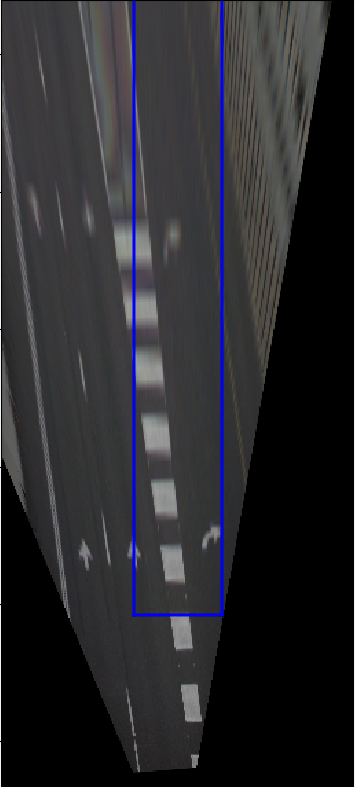}  \end{tabular}
\caption{\textbf{Self-supervision task for lane detection.} Depicted is the same image transformed for the right, center and left viewing directions. The network then gets the central rectangle (outlined in blue) and has to predict the viewing angle it was generated with.}
\label{fig:self}
\end{figure}

\subsection{Autoencoder approach}\label{sec:AE}
Our proposed autoencoder approach is inspired by the human landmark detection approach of~\cite{jakab2019learning}, in which a landmark detector is trained with unlabeled target domain images, and unpaired ground truth annotations. In our application, this approach is based on the task-specific assumption that in the road area lane markings correlate with image gradients. Essentially, we train a detector to generate, from an unlabeled input image $x$, a gray-scale ``lane image'' that satisfies two constraints: (1) it looks like a valid ground truth image of lanes and (2) it holds the information required to reconstruct the original gradients in the image. The first constraint is imposed by a discriminator, $\gamma$, which tries to distinguish between the generated lane image $l$ and unpaired ground truth lane annotations. The second is imposed by a decoder that tries to reconstruct the original image gradients $Grad(x)$ from $l$. The entire approach, illustrated in Figure~\ref{fig:arch}(b), relies on the assumption, that the natural candidate for the encoded image $l$ satisfying both constraints is that of the lanes in the scene. 

Formally, our training process gets as input the unlabeled target domain images $\mathcal{D}^u_T$ and a set of source domain ground truth ``lane images'', $\mathcal{D}^{up}_S=\{\overline{l}\}$, as described next. Given an input image $x\in{D}^u_T$ and its corresponding feature maps $f=\phi(x)$, an encoder, $\delta:\mathcal{F}\mapsto \mathcal{I}_L=\mathbb{R}^{W/4,H/4}$, generates a gray-scale ``lane image'' $l=\delta(f)\in\mathcal{I}_L$. The discriminator, $\gamma$, is trained to distinguish between the generated lane images and the ground truth lane images. The decoder $\lambda$, generates an image $g$, trained to reconstruct $Grad(x)$ using an $L_1$ loss:
\begin{equation*}
    \mathcal{L}_{reconstruct}=\sum_{x\in \mathcal{D}_T^u}\|\lambda \circ \delta \circ \phi(x)-Grad(x)\|_{1}
\end{equation*} 

The GAN loss functions consist of the discriminator loss function $\mathcal{L}_d$, minimized for the discriminator ($\gamma$) parameters, and the generator loss function $\mathcal{L}_g$, minimized for the parameters of the generator which in our case is $\delta\circ\phi$. We chose $\mathcal{L}_g, \mathcal{L}_d$ following the Wasserstein GAN with gradient penalty (WGAN-GP) suggested by~\cite{Gulrajani:nips:2017:wgangp}. Training the network alternates between two objectives: minimizing the discriminator loss $\mathcal{L}_d$ over the parameters of the discriminator, $\gamma$, and minimizing a combination of the reconstruction and generator losses $\alpha L_{reconstruct}+ L_g$ over the parameters of $\delta$, $\phi$ and $\lambda$. 

\

One hurdle to overcome in this approach is the simple fact that not all gradients in the top-view image $Grad(x)$ can be explained by the lane image, with edges belonging to other road markings, on-road objects, sidewalks and even the lane mark dashing themselves. We implement several heuristics to mitigate the effect of all other factors: we apply a car detector (YOLOv3~\cite{redmon2018yolov3}) and mark the detected pixels to be ignored in the reconstruction loss. Similarly, we ignore everything outside a dilated convex hull around the lanes detected in $l$. Finally, as can be seen in Figure~\ref{fig:arch}(b), the decoder, $\lambda$, manages to reconstruct other road markings and lane dash positions, by hiding appearance ques within $l$, a phenomena observed by~\cite{jakab2019learning} as well. This base architecture can be trained in an \textbf{unpaired labels} setting, i.e. without having any source domain images, but with only the set of unpaired ground truth lane images from the source domain, $\mathcal{D}^{up}_S$. While in this setting the method converged at times (See result example in the appendix), in general it is highly unstable. This instability was solved by adding direct task supervision using the source dataset in the UDA setting as described next.

In the UDA setting we additionally have source domain images paired with their labels, $\mathcal{D}^l_S$. The data from source domain has, thus, two roles while training - pairs of images and labels are used  to minimize $\mathcal{L}_{task}$, and labels are also used to generate lane images $\overline{l}$ for the autoencoder adversarial learning. Figure~\ref{fig:da_arch}(a) illustrates this UDA architecture. Notice that the feature embedding, $\phi$, is shared between the two domains, and hence is the network in which domain adaptation occurs. On the other hand, the high level reasoning network ($\psi$) producing the final lane result, sees only source domain images, and hence expects the feature maps $f$ to be domain invariant. We follow this decomposition for all our tested domain adaptation methods.

\subsection{Self-supervision by viewing orientation prediction}\label{sec:Self}

In self-supervision it is important to find an auxiliary task that will train a representation informative for the end-goal task. To this end, we introduce a new self-supervision task illustrated in Figure~\ref{fig:self}. The basic idea is to generate one of three viewing angles of the scene and let the network predict which one it is. Since the vehicle orientation is commonly aligned with the lanes, predicting the camera viewing direction is strongly related to the capability to correctly detect the lanes. Formally, each un-warped image $I$ is transformed to a virtual top view according to a selected orientation $\hat{o}$: either the regular central one ($c$), or one of two additional views in which the camera is virtually panned by $5$ degrees to the left ($l$) or to the right ($r$). We then crop from the resulting top-view a rectangular image from a fixed area in the center of the view. This orientation dependent transformation, $\eta$, is equivalent, up to image sampling differences, to cropping an oriented rectangle from the top-view image $x$. The self-supervision task is then to predict the orientation $o$ from the cropped rectangle $I_r$. To this end we train a  small classifier network $\rho$ on top of the embedding features: $o=\rho\circ\phi\circ\eta(I,\hat{o})$.  The self-supervision objective function $\mathcal{L}_{self}$ is the cross-entropy loss for the three-way classification task. In the UDA setting we train using the self-supervision objective by minimizing $\mathcal{L}_{self}$ for the unlabeled target images and $\mathcal{L}_{task}$ for the labeled source images, as illustrated in Figure~\ref{fig:da_arch}.

\subsection{Image translation}\label{sec:IT}

As proposed in the Cycada framework for UDA~\cite{Hoffman:arxiv:2017:Cycada}, we use CycleGAN~\cite{Zhu:iccv:2017:cycle} for image-to-image translation. Figure~\ref{fig:da_arch}(c) illustrates the approach. Source images $x_s$ are mapped to target images $x_t$. We slightly abuse notations for simplicity because here $x_s,x_t$ are the original unwarped images, not in top-view. A discriminator  $\gamma_t$ and generator $G_{s\rightarrow t}$ are trained using the loss:
\begin{align}
    \mathcal{L}_{\text{GAN}}(G_{s\rightarrow t},\gamma_t) &= \min_{G_{s\rightarrow t}}\max_{\gamma_t}\left(\sum_{x_t\in \mathcal{D}^u_t} [\log \gamma_t\left(x_t\right)] + 
\sum_{x_s\in \mathcal{D}^l_s}[\log (1-\gamma_t\left(G_{s\rightarrow t}\left(x_s\right)\right)]\right) \nonumber
\end{align}
A discriminator $\gamma_s$ and generator $G_{t\rightarrow s}$ are trained analogously.

As in CycleGAN~\cite{Zhu:iccv:2017:cycle}, cycle-consistency loss is used to ensure image content is preserved while translating the original image.
This allows using the source labels $y_s$ with the translated source images, $G_{s\rightarrow t}(x_s)$, to minimize the task loss, $\mathcal{L}_{task}$. We  optimize the task loss in a second stage over the translated images (transformed to top-view). Combined single-stage training of the CycleGAN and task loss did not improve performance. We also tried to combine feature-level distribution matching as proposed in the Cycada framework~\cite{Hoffman:arxiv:2017:Cycada}, but did not gain any accuracy improvement.

\subsection{Feature-level distribution matching}\label{sec:DM}
We chose implementing a GAN approach~\cite{Ganin_ICML15} and the Central Moment Discrepancy CMD~\cite{Zellinger:arxiv:2017:cmd} for feature-level distribution matching. The goal is to push the distribution of the feature map representations for the source domain ($f_s$) and the target domain ($f_t$) towards one another. In the \textbf{embedding GAN} framework, we train a discriminator to distinguish between the domain of different examples from the feature embedding and a generator, in our case the feature embedding function $\phi$, to fool the discriminator. Again, we use the GAN loss functions $\mathcal{L}_g, \mathcal{L}_d$ following the WGAN-GP formulation~\cite{Gulrajani:nips:2017:wgangp}. In the UDA setting, $\mathcal{L}_{task}$ and $\mathcal{L}_g$ are simultaneously minimized over the source images. Figure~\ref{fig:da_arch}(d) illustrates this approach. In the \textbf{CMD} approach the GAN is replaced by a distance loss on the first two central moments between the two distributions as in~\cite{Zellinger:arxiv:2017:cmd}. Details for the latter approach are brought in the appendix.

\section{Experiments}\label{sec:experiments}
We test the various approaches for domain adaptation on three different lane detection benchmarks. In each experiment, we set the synthetic dataset as the source domain, and one of the three benchmarks as the target domain. Each training session is run for a fixed number of iterations. 
In the semi-supervised setting, we choose a \textbf{labeled subset} of the training data (roughly 10\%), consisting of one or several, separate driving sessions.

\subsection{Datasets}
\textbf{Synthetic dataset}. In all our experiments, we follow the methodology proposed in~\cite{Garnett:cvpr:2019:3d} to generate a synthetic dataset as the source domain. Images are generated by randomly drawing the parameters for each scene using the generation method  from~\cite{Garnett:cvpr:2019:3d} (See examples in Figure~\ref{fig:intro}(a)). What is unique about this methodology is its simplicity. As opposed to manually or semi-manually generated synthetic datasets such as~\cite{vKITTI,synthia}, this methodology uses a very small set of graphical assets and instead achieves scene variability by randomly varying lane topology and geometry. In this sense, it can be viewed as "free data" which can be generated using a simple algorithm and an open-source graphic engine. For each target domain, we generated 50K examples using the same scene parameters, but with camera intrinsic and extrinsic parameters roughly adjusted to the target domain. In our experiments generating more images did not result in accuracy gains. This may be due to the limited variability of the appearance in our synthetic data, or due to the limited variability of the lane geometry in the target domains. 

\textbf{tuSimple}. The tuSimple lane dataset~\cite{tuSimple} consists of 3,626 training and 2,782 test images in mostly highway scenarios. While relatively small and not very diverse, we chose it for being the most studied. We report results on the test set.

\textbf{llamas}. The llamas lane-marker dataset~\cite{Nehrendt:ICCV:2019:llamas} is a newer, much larger dataset, consisting of over 100K labeled images. Created from 14 highway recordings of around 25 km each, together with high accuracy maps in a fully automatic process, it is one of the largest datasets available today. As labels are not available for the test set, we show the evaluation results on a validation set, which consists of a driving session not used during training.

\textbf{3DLanes}. The 3DLanes~\cite{netalee} dataset is the most recent and most diverse dataset among the three. It contains 330000 images labeled in a semi-automated process, in highway and rural environments. Again, we report results on the validation set.

\subsection{Evaluation Metrics}

\textbf{Segment-based evaluation}. In~\cite{netalee}, evaluation is preformed only after lane-level clustering. In this study,  the goal is to compare the different DA methods directly, and therefore we refrain from applying post-processing methods that can skew the results. For this purpose, we propose a direct tile-representation evaluation formulation. The segment-based evaluation follows the principles of object detection evaluation, where detected lane segments are matched to ground truth ones, and Precision-Recall curves are computed for different matching criteria. 

For a single image, the input for the evaluation is an unordered set $SEG_{out}$ comprising of all output lane segments $\mathbf{p}$ (with corresponding detection confidence score $b$),  and the set of all ground truth lane segments obtained using the same representation $SEG_{gt}=\{\mathbf{\hat{p}}\}$. We first compute a symmetric lane segment distance $seg\_dist(\mathbf{p},\mathbf{\hat{p}})$ for all segment pairs ($\mathbf{p}\in SEG_{out},\mathbf{\hat{p}}\in SEG_{gt}$), reflecting a geometric distance between the segments in the top view (see appendix for a detailed formulation of $seg\_dist$). We then use the Hungarian algorithm~\cite{Kuhn:naval:1955:hungarian} to find a minimum distance matching between the two sets of lane segments. Given a maximum distance $seg\_dist_{max}$ (analogous to maximum IoU in object detection), We compute the Recall-Precision (RP) curve, considering all segment matches (in all the test images) with $seg\_dist<seg\_dist_{max}$, by iterating over the confidence values ($b$) of the detected segments. Each RP curve is summarized as the Average Precision (AP). Our final evaluation metric, the mean Average Precision (mAP), is computed as the the average AP for five different maximum matching distances: $seg\_dist_{max}\in\{10cm,20cm,30cm,40cm,50cm\}$. These distances correspond to real-world distances in the road plane.

\textbf{Lane-based evaluation.} For completeness, we also compute the lane-based evaluation metric used in the tuSimple benchmark~\cite{tuSimple}. This metric requires a single detection per lane, and hence further post-processing, namely clustering of output tiles. To this end, we deploy a simple, heuristic clustering algorithm described in the appendix. In this study, there are several disadvantages to using it. A partial list includes:
\begin{itemize}
    \item  it is indirect by relying on an additional component - the clustering
    \item it evaluates performance in the image plane and not in top view as our method outputs
    \item  it assumes a known number of lanes in the scene
    \item we observed that in practice that it is less correlated with performance of the detector
\end{itemize}

\subsection{Implementation details}
All the implemented modules are convolution neural networks. The feature embedding function, $\phi$, is based on the VGG architecture~\cite{VGG}. All experiments were run from random initialization for a pre-defined number of iterations. We use the ADAM optimizer~\cite{adam} without weight decay. For most methods we observed a non-negligible variability in performance between training iterations, even towards the end of each training session. To reduce the effect of this phenomena, all the results we report are averaged over 5 different snapshots 100 iterations apart at the end of the session. The remaining architecture and training protocol details are provided in the appendix.

\subsection{Results}

Table~\ref{tab:res} summarizes our results on each of the three datasets. The first row shows the \textbf{fully supervised} result, which serves as the upper bound for all other tested methods. Compared with state-of-the-art (96.6\%,~\cite{Hou:ICCV:2019:learning}) our base supervised method without any bells and whistles reaches 95.1\% on the tuSimple benchmark using the tuSimple lane-based evaluation metrics. All the results in Table~\ref{tab:res} use our mAP metric. Apparent from the experiments is that the llamas and 3DLanes datasets are indeed more difficult compared to the tuSimple. The \textbf{synthetic only} model is trained with the synthetic labeled data without performing any adaptation to the target domain. Of all tested methods, it gives the poorest results due to the noticeable, significant domain gap that can be observed in Figure~\ref{fig:intro}. This model serves as the baseline for all UDA methods. 

\textbf{Unsupervised domain adaptation}. In the UDA setting,  The baseline-gap (BG) column, specifies the portion of the accuracy gap between the \textbf{fully supervised} and the \textbf{synthetic only} models, closed by the corresponding method. From the five domain adaptation method tested, three methods, namely \textbf{self-supervision (S)}, \textbf{image translation (IT)} and \textbf{autoencoder (AE)}, gained significant improvements over the non-adapted baseline (\textbf{synthetic only}). The remaining two methods, \textbf{CMD} and \textbf{embedding GAN} gave worse results in most experiments. Among the three leading methods, performance is similar, with slight advantages to one over the other in different experiments. We also tried all combinations of these three methods, with some combinations bringing small additional improvements. Details on the training strategy for combined domain adaptation methods are in the appendix. On the tuSimple and 3DLanes benchmarks the \textbf{image-translation} method was the single best performing method while on llamas \textbf{self-supervision} performed best. different combinations of the three leading methods seem to further improve performance, and combining all three (\textbf{AE+IT+S}), performed best on two out of the three datasets closing more than 70\% of the baseline-gap. Notably, in all experiments, training with self-supervision was more stable and reproducible compared with the two competing methods that rely on adversarial training.

\textbf{Semi-supervised.}
In this setting we have roughly 10\% of the target domain labels. We start with the simplest option, \textbf{small supervision}, in which we train a supervised model using this small portion of target domain data. Interestingly, adding labeled synthetic data (\textbf{small+syn. supervision}) to the train set, significantly improves performance for all three datasets even without domain adaptation. The latter serves as the baseline for all semi-supervised domain adaptation methods, and used to compute the $BG$ measure. In contrast with the other methods, self-supervision does not require source domain data, but can exploit unlabeled target domain images. Our experiments show that this method, \textbf{self-sup. w/o syn}, improves accuracy in two out of the three benchmarks.

\textbf{Semi-supervised domain adaptation}
As in the UDA setting, we evaluated all domain adaptation methods and combinations, but this time with an extra portion of labeled target domain data. As in the UDA setting we observed inferior performance for the \textbf{CMD} and the \textbf{embedding GAN} methods, omitted in Table~\ref{tab:res} for brevity. As expected the additional labeled data improves the resulting accuracy for each method compared to the UDA setting. Our proposed autoencoder approach outperforms all other single methods on all datasets and all method combinations except for one - \textbf{AE+S} on llamas. On tuSimple and llamas the autoencoder approach respectively delivers 1.8\% and 2.7\% mAP less than the fully supervised model. The practical implication of this result is that using our approach, sacrificing this small loss in accuracy provides a ten fold saving in data annotation. We also note that each experiment we present requires a significant computational effort, and therefore a systematic study of semi-supervision with different amounts of labeling is beyond the scope of this study.

\begin{table}
    \centering
    \begin{tabular}{cccccccccc}
        \Xhline{4\arrayrulewidth}
         & & & & \multicolumn{2}{c}{\underline{tuSimple}} & \multicolumn{2}{c}{\underline{llamas}} & \multicolumn{2}{c}{\underline{3DLanes}}\\
         & & {\%lbl.} & {Syn.} &  \multirow{2}{*}{{mAP\%}}  & \multirow{2}{*}{{BG\%}}  &
          \multirow{2}{*}{{mAP\%}}  &  \multirow{2}{*}{{BG\%}}  & 
          \multirow{2}{*}{{mAP\%}}  &  \multirow{2}{*}{{BG\%}} \\
          &  & {data} & {data} & & & & & &  \\
        \Xhline{4\arrayrulewidth}

        & {fully supervised} & 100 & - & 81.1 &  100 & 73.2 & 100 & 74.5 & 100\\
        & {synthetic only} & - & \checkmark & 60.5    & 0 & 47.8 &    0 & 53.8     & 0 \\
        \hline
        \multirow{9}{*}{\rotatebox[origin=c]{90}{\textbf{UDA}}} 
        & {Autoencoder}   & - & \checkmark & 67.7 & 35.0 & 56.0 & 32.3 & 57.8 & 19.1  \\
        & {Image translation} & - & \checkmark & \textbf{72.0} & \textbf{55.7} & 62.3 & 57.3 & \textbf{59.3} & \textbf{26.5}  \\
        & {Self supervision}  & - & \checkmark & 66.3 & 27.9 & \textbf{63.4} & \textbf{61.6} & 58.1 & 20.5 \\
        & {CMD}   & - & \checkmark & 62.7 & 10.6 & 51.9 & 15.9 & 55.7 & 9.0 \\
        & {Embedding GAN}     & - & \checkmark & 52.1 & 0 & 37.8 & 0    & 31.9 & 0    \\
        & {AE+IT}   & - & \checkmark & 73.4 & 62.5 & 65.6 & 70.0 & 59.6 & 27.8 \\
        & {AE+S}    & - & \checkmark & 70.1 & 46.3 & 63.7 & 62.6 & 59.7  & 28.7 \\
        & {IT+S}    & - & \checkmark & 72.4 & 57.8 & 65.3 & 69.1 & \textbf{60.1} & \textbf{30.6} \\
        & {AE+IT+S} & - & \checkmark & \textbf{77.1} & \textbf{80.5} & \textbf{66.0} & \textbf{71.6} & 59.5 & 27.5 \\
        \hline
        & {small supervision} & 10 & - & 74.4 & - & 60.1 & - & 56.9 & -  \\
        & {self-sup. w/o syn} & 10 & - & 75.2 & - & 68.6 & - & 60.1 &  - \\
        & {small+syn. supervision} & 10 & \checkmark & 77.5
 & 0 & 65.5 & 0 & 62.0 & 0  \\
        \hline
        \multirow{9}{*}{\rotatebox[origin=c]{90}{\textbf{SSDA}}} 
        & {Autoencoder}   & 10 & \checkmark & \textbf{79.3} & \textbf{50.0} & \textbf{70.5} & \textbf{65.6} & \textbf{66.9} & \textbf{39.0} \\
        & {Image translation} & 10 & \checkmark & 78.4 &  25.0 & 69.1 & 47.1 & 64.2 & 17.5  \\
        & {Self supervision}  & 10 & \checkmark & 77.1 &    0 & 70.2 & 61.1 & 64.0 & 16.0 \\
        & {AE+IT}             & 10 & \checkmark & 78.8 & 37.7 & 70.3 & 63.4 & 66.2 & 33.3 \\
        & {AE+S}              & 10 & \checkmark & 77.6 & 4.6 & \textbf{70.7} & \textbf{68.3} & 65.2 & 25.6  \\
        & {IT+S}              & 10 & \checkmark & 77.8 & 8.9 & 70.1 & 59.6 & 65.8 & 30.4 \\
        & {AE+IT+S}           & 10 & \checkmark & 77.7  & 5.2 & 70.4 & 64.5 & 66.6 & 36.7   \\
        \Xhline{4\arrayrulewidth}
    \end{tabular}
    \caption{Results on the tuSimple~\cite{tuSimple}, llamas~\cite{Nehrendt:ICCV:2019:llamas} and 3DLanes~\cite{Garnett:cvpr:2019:3d} datasets using the mean Average Precision (mAP) using the \textbf{segment-based} evaluation. Values in the column  ``\textbf{\%lbl. data}'' correspond to the percent of labeled data \textit{from the respective target domain} used by each method. The ``\textbf{Syn. data}'' column specifies whether synthetic data was used in training. \textbf{Unsupervised domain adaptation:} rows in the section marked \textbf{UDA} correspond to unsupervised domain adaptation experiments. The BG\% (Baseline Gap) for these methods measures the percent of the gap closed between the \textbf{synthetic only} and the \textbf{fully supervised} mAP: BG(\textbf{method})=(mAP(\textbf{method})-mAP(\textbf{synthetic only}))/(mAP(\textbf{fully supervised})-mAP(\textbf{synthetic only})). The best single method and best method combination are marked in bold. \textbf{Semi-supervised domain adaptation:} rows in the section marked \textbf{SSDA} correspond to semi-supervised domain adaptation experiments. The BG\% (Baseline Gap) for these methods measures the percent of the gap closed between the \textbf{small+syn. supervision} and the \textbf{fully supervised} mAP: BG(\textbf{method})=(mAP(\textbf{method})-mAP(\textbf{small+syn. supervision}))/(mAP(\textbf{fully supervised})-mAP(\textbf{small+syn. supervision})). The best single method and best method combination are marked in bold. }
    \label{tab:res}
\end{table}

\section{Conclusions}

We showed that it is possible to improve lane detection when labeled data availability is limited using domain adaptation from random synthetic data. To this end we introduced a new autoencoder approach for domain adaptation and a new task-specific self-supervision objective. We further explored the different existing domain adaptation approaches and their effectiveness for the lane detection task. While these findings have practical implications in autonomous driving research, we suggest further study beyond the scope of this work: experimenting with varying amounts of supervised data in the semi-supervised setting, exploring the effect of the complexity of the synthetic data on the final results, and adapting the detector for lanes in \textit{3D}.
\clearpage
\bibliographystyle{splncs}
\bibliography{egbib}

\begin{thebibliography}{10}

\bibitem{Hou:ICCV:2019:learning}
Hou, Y., Ma, Z., Liu, C., Loy, C.C.:
\newblock Learning lightweight lane detection cnns by self attention
  distillation.
\newblock In: ICCV. (2019)  1013--1021

\bibitem{Garnett:cvpr:2019:3d}
Garnett, N., Cohen, R., Pe'er, T., Lahav, R., Levi, D.:
\newblock 3d-lanenet: end-to-end 3d multiple lane detection.
\newblock In: ICCV. (2019)  2921--2930

\bibitem{tuSimple}
:
\newblock
\newblock (\texttt{http://benchmark.tusimple.ai}, lane challenge)

\bibitem{waymo_dataset}
Sun, P., Kretzschmar, H., Dotiwalla, X., Chouard, A., Patnaik, V., Tsui, P.,
  Guo, J., Zhou, Y., Chai, Y., Caine, B., Vasudevan, V., Han, W., Ngiam, J.,
  Zhao, H., Timofeev, A., Ettinger, S., Krivokon, M., Gao, A., Joshi, A.,
  Zhang, Y., Shlens, J., Chen, Z., Anguelov, D.:
\newblock Scalability in perception for autonomous driving: Waymo open dataset.
\newblock CoRR \textbf{abs/1912.04838} (2019)

\bibitem{Cesar:arxiv:2019:nuscenes}
Caesar, H., Bankiti, V., Lang, A.H., Vora, S., Liong, V.E., Xu, Q., Krishnan,
  A., Pan, Y., Baldan, G., Beijbom, O.:
\newblock nuscenes: A multimodal dataset for autonomous driving.
\newblock arXiv preprint arXiv:1903.11027 (2019)

\bibitem{Gaidon:cvpr:2016:virtual}
Gaidon, A., Wang, Q., Cabon, Y., Vig, E.:
\newblock Virtual worlds as proxy for multi-object tracking analysis.
\newblock In: CVPR. (2016)  4340--4349

\bibitem{vKITTI}
Gaidon, A., Wang, Q., Cabon, Y., Vig, E.:
\newblock Virtualworlds as proxy for multi-object tracking analysis.
\newblock In: CVPR, {IEEE} Computer Society (2016)  4340--4349

\bibitem{jakab2019learning}
Jakab, T., Gupta, A., Bilen, H., Vedaldi, A.:
\newblock Learning landmarks from unaligned data using image translation.
\newblock CoRR \textbf{abs/1907.02055} (2019)

\bibitem{Hoffman:arxiv:2017:Cycada}
Hoffman, J., Tzeng, E., Park, T., Zhu, J.Y., Isola, P., Saenko, K., Efros,
  A.A., Darrell, T.:
\newblock Cycada: Cycle consistent adversarial domain adaptation.
\newblock In: ICML. (2018)

\bibitem{Ganin_ICML15}
Ganin, Y., Lempitsky, V.:
\newblock Unsupervised domain adaptation by backpropagation.
\newblock In: ICML. Volume~37 of ICML’15., JMLR.org (2015)  1180–1189

\bibitem{Zellinger:arxiv:2017:cmd}
Zellinger, W., Grubinger, T., Lughofer, E., Natschl{\"{a}}ger, T.,
  Saminger{-}Platz, S.:
\newblock Central moment discrepancy {(CMD)} for domain-invariant
  representation learning.
\newblock In: 5th International Conference on Learning Representations, {ICLR}
  2017, Toulon, France, April 24-26, 2017, Conference Track Proceedings,
  OpenReview.net (2017)

\bibitem{Nehrendt:ICCV:2019:llamas}
Behrendt, K., Soussan, R.:
\newblock Unsupervised labeled lane markers using maps.
\newblock In: ICCV. (2019)

\bibitem{netalee}
Efrat, N., Bluvstein, M., Garnett, N., Levi, D., Oron, S., Shlomo, B.E.:
\newblock Semi-local 3d lane detection and uncertainty estimation.
\newblock CoRR \textbf{abs/2003.05257} (2020)

\bibitem{Bousmalis_NIPS2016}
Bousmalis, K., Trigeorgis, G., Silberman, N., Krishnan, D., Erhan, D.:
\newblock Domain separation networks.
\newblock In: NIPS.
\newblock Curran Associates, Inc. (2016)  343--351

\bibitem{Long_ICML15}
Long, M., Cao, Y., Wang, J., Jordan, M.I.:
\newblock Learning transferable features with deep adaptation networks.
\newblock In: ICML. Volume~37 of ICML’15., JMLR.org (2015)  97–105

\bibitem{Adda_CVPR2017}
Tzeng, E., Hoffman, J., Darrell, T., Saenko, K.:
\newblock Adversarial discriminative domain adaptation.
\newblock In: CVPR. (2017)

\bibitem{Long_NIPS2018_7436}
Long, M., CAO, Z., Wang, J., Jordan, M.I.:
\newblock Conditional adversarial domain adaptation.
\newblock In Bengio, S., Wallach, H., Larochelle, H., Grauman, K.,
  Cesa-Bianchi, N., Garnett, R., eds.: NIPS.
\newblock (2018)  1640--1650

\bibitem{lahiri2018unsupervised}
Lahiri, A., Agarwalla, A., Biswas, P.K.:
\newblock Unsupervised domain adaptation for learning eye gaze from a million
  synthetic images: An adversarial approach.
\newblock CoRR \textbf{abs/1810.07926} (2018)

\bibitem{Gen2Adapt}
Sankaranarayanan, S., Balaji, Y., Castillo, C.D., Chellappa, R.:
\newblock Generate to adapt: Aligning domains using generative adversarial
  networks.
\newblock CoRR \textbf{abs/1704.01705} (2017)

\bibitem{Zhu:iccv:2017:cycle}
Zhu, J.Y., Park, T., Isola, P., Efros, A.A.:
\newblock Unpaired image-to-image translation using cycle-consistent
  adversarial networks.
\newblock In: ICCV. (2017)  2223--2232

\bibitem{Huang:iccv:2017:adain}
Huang, X., Belongie, S.:
\newblock Arbitrary style transfer in real-time with adaptive instance
  normalization.
\newblock In: ICCV. (2017)  1501--1510

\bibitem{TaigmanPW17}
Taigman, Y., Polyak, A., Wolf, L.:
\newblock Unsupervised cross-domain image generation.
\newblock In: ICLR, OpenReview.net (2017)

\bibitem{BousmalisSDEK17}
Bousmalis, K., Silberman, N., Dohan, D., Erhan, D., Krishnan, D.:
\newblock Unsupervised pixel-level domain adaptation with generative
  adversarial networks.
\newblock In: CVPR, IEEE Computer Society (2017)  95--104

\bibitem{CrDoCo}
Chen, Y.C., Lin, Y.Y., Yang, M.H., Huang, J.B.:
\newblock Crdoco: Pixel-level domain transfer with cross-domain consistency.
\newblock (2019)

\bibitem{NorooziF16}
Noroozi, M., Favaro, P.:
\newblock Unsupervised learning of visual representations by solving jigsaw
  puzzles.
\newblock In: ECCV, Springer (2016)  69--84

\bibitem{GidarisSK18}
Gidaris, S., Singh, P., Komodakis, N.:
\newblock Unsupervised representation learning by predicting image rotations.
\newblock In: ICLR. (2018)

\bibitem{oord2018representation}
van~den Oord, A., Li, Y., Vinyals, O.:
\newblock Representation learning with contrastive predictive coding.
\newblock CoRR \textbf{abs/1807.03748} (2018)

\bibitem{hnaff2019dataefficient}
H{\'{e}}naff, O.J., Razavi, A., Doersch, C., Eslami, S.M.A., van~den Oord, A.:
\newblock Data-efficient image recognition with contrastive predictive coding.
\newblock CoRR \textbf{abs/1905.09272} (2019)

\bibitem{sun2019unsupervised}
Sun, Y., Tzeng, E., Darrell, T., Efros, A.A.:
\newblock Unsupervised domain adaptation through self-supervision.
\newblock CoRR \textbf{abs/1909.11825} (2019)

\bibitem{chen2018domain}
Chen, Y., Li, W., Sakaridis, C., Dai, D., Van~Gool, L.:
\newblock Domain adaptive faster r-cnn for object detection in the wild.
\newblock In: CVPR. (2018)

\bibitem{zhu2019adapting}
Zhu, X., Pang, J., Yang, C., Shi, J., Lin, D.:
\newblock Adapting object detectors via selective cross-domain alignment.
\newblock In: CVPR. (2019)  687--696

\bibitem{AbarghoueiB18}
Abarghouei, A.A., Breckon, T.P.:
\newblock Real-time monocular depth estimation using synthetic data with domain
  adaptation via image style transfer.
\newblock In: CVPR 2018, {IEEE} Computer Society (2018)  2800--2810

\bibitem{zheng2018t2net}
Zheng, C., Cham, T.J., Cai, J.:
\newblock T2net: Synthetic-to-realistic translation for solving single-image
  depth estimation tasks.
\newblock In: ECCV. (2018)  767--783

\bibitem{Hong_CVPR18}
{Hong}, W., {Wang}, Z., {Yang}, M., {Yuan}, J.:
\newblock Conditional generative adversarial network for structured domain
  adaptation.
\newblock In: CVPR. (2018)  1335--1344

\bibitem{Zhang_2018}
Zhang, Y., Qiu, Z., Yao, T., Liu, D., Mei, T.:
\newblock Fully convolutional adaptation networks for semantic segmentation.
\newblock Proceedings of CVPR (2018)

\bibitem{Sankaranarayanan18}
Sankaranarayanan, S., Balaji, Y., Jain, A., Lim, S., Chellappa, R.:
\newblock Learning from synthetic data: Addressing domain shift for semantic
  segmentation.
\newblock In: CVPR, {IEEE} Computer Society (2018)  3752--3761

\bibitem{Chen_CVPR19}
{Chen}, Y., {Li}, W., {Chen}, X., {Van Gool}, L.:
\newblock Learning semantic segmentation from synthetic data: A geometrically
  guided input-output adaptation approach.
\newblock In: CVPR. (2019)  1841--1850

\bibitem{synthia}
{Ros}, G., {Sellart}, L., {Materzynska}, J., {Vazquez}, D., {Lopez}, A.M.:
\newblock The synthia dataset: A large collection of synthetic images for
  semantic segmentation of urban scenes.
\newblock In: CVPR. (2016)  3234--3243

\bibitem{Gulrajani:nips:2017:wgangp}
Gulrajani, I., Ahmed, F., Arjovsky, M., Dumoulin, V., Courville, A.C.:
\newblock Improved training of wasserstein gans.
\newblock In: NIPS. (2017)  5767--5777

\bibitem{redmon2018yolov3}
Redmon, J., Farhadi, A.:
\newblock Yolov3: An incremental improvement.
\newblock CoRR \textbf{abs/1804.02767} (2018)

\bibitem{Kuhn:naval:1955:hungarian}
Kuhn, H.W.:
\newblock The hungarian method for the assignment problem.
\newblock Naval research logistics quarterly \textbf{2} (1955)  83--97

\bibitem{VGG}
Simonyan, K., Zisserman, A.:
\newblock Very deep convolutional networks for large-scale image recognition.
\newblock CoRR \textbf{abs/1409.1556} (2014)

\bibitem{adam}
Kingma, D.P., Ba, J.:
\newblock Adam: A method for stochastic optimization.
\newblock arXiv preprint arXiv:1412.6980 (2014)

\bibitem{miyato2018spectral}
Miyato, T., Kataoka, T., Koyama, M., Yoshida, Y.:
\newblock Spectral normalization for generative adversarial networks.
\newblock In: International Conference on Learning Representations. (2018)

\bibitem{SalimansGZCRC16}
Salimans, T., Goodfellow, I.J., Zaremba, W., Cheung, V., Radford, A., Chen, X.:
\newblock Improved techniques for training gans.
\newblock CoRR \textbf{abs/1606.03498} (2016)

\end{thebibliography}
\nocite{miyato2018spectral}
\nocite{SalimansGZCRC16}
\newpage
\appendix
\section{Appendix - implementation details}

\subsection{Training}
All training sessions consisted of 30,500 iterations (batches) with constant learning rate and no weight decay. In case of adversarial training, on each iteration, both discriminator and generator were updated. Test results reported were averaged over results obtained with model snapshots at iterations 30,100, 30,200, 30,300, 30,400 and 30,500.

\textbf{Batch size.} For all supervised methods (optimizing $\mathcal{L}_{task}$), including the supervision over translated images in the \textbf{image translation} method, we used batch size 24. In cases in which supervised training is with two domains (e.g. \textbf{small + syn. supervision}) each batch consisted of 12 examples from each domain. For the other domain adaptation methods, \textbf{CMD}, \textbf{autoencoder}, \textbf{embedding GAN} and \textbf{self-supervision} methods, each batch consisted of 16 source domain images and 16 target ones. 

\textbf{learning rate.} For all weight updates the learning is $10^{-4}$ except for the discriminator weights updated with learning rate $5\cdot 10^{-4}$.

\textbf{Loss balancing.} For all wgan-gp implementation the gradient penalty loss weight was 10. In adversarial training the generator loss weight is 0.2. For the \textbf{CMD} the regularization loss weight is $10^{-3}$. For the autoencoder the weight of the reconstruction loss $\alpha=10$.

\textbf{DA method combination}. Whenever DA methods are combined (e.g. \textbf{AE+S}), the same target domain batch is used multiple times (one per method), each time for optimizing a different loss.

\subsection{CMD}
Moment matching aims at matching the distributions of the intermediate feature maps of the source and target domain by minimizing the distance between their first order moments. We follow the method described in~\cite{Zellinger:arxiv:2017:cmd}. Given a batch of source and domain examples their corresponding feature maps $f_s,f_t$ with $C$ channels in each are computed. For each example, each spatial entry in the feature map is considered as a single $C$ dimensional sample. The collection of all samples from each domain $z_s,z_t\in\mathbb{R}^C$ is then the input to the central moment discrepancy loss~\cite{Zellinger:arxiv:2017:cmd}.

\subsection{Evaluation}

\begin{figure}[ht]
    \centering
    \begin{tabular}{c|c}
\includegraphics[height=4cm]{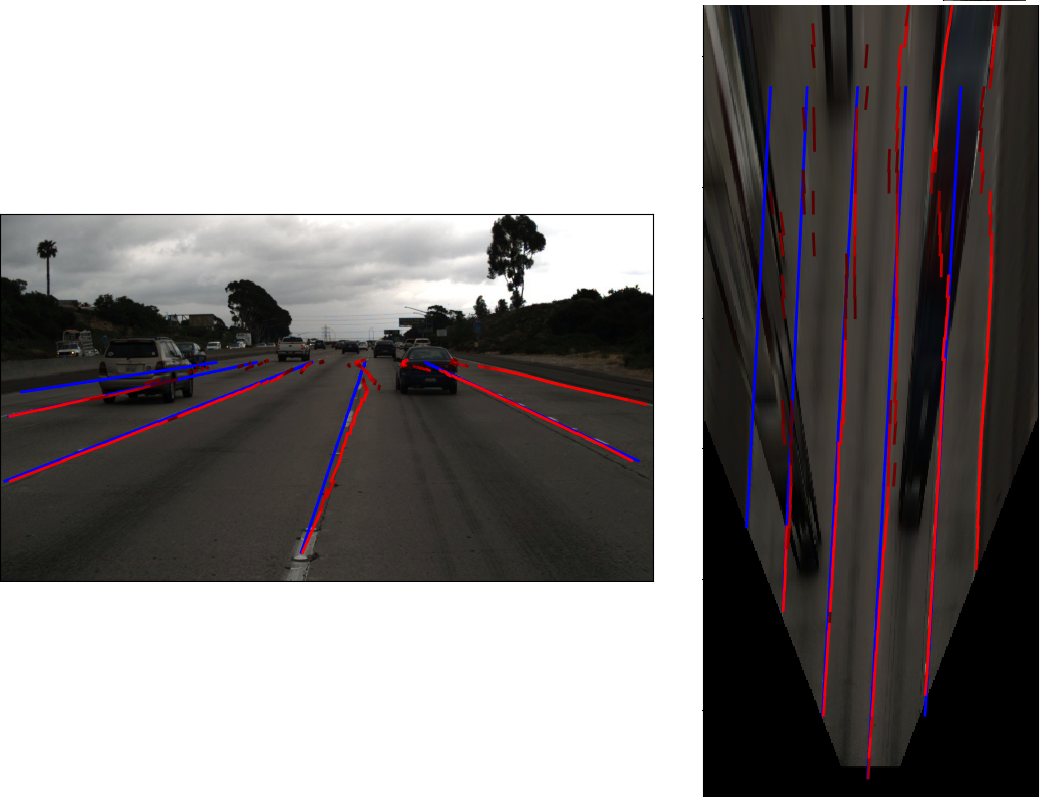}& 
         \includegraphics[height=4cm]{figures/tusimple_0_931.png}
    \end{tabular}
    
    \caption{Two examples of the forgivingness of the tuSimple evaluation towards the lane segment output. Ground truth marked in blue and detected segments in red. In both examples the per-image tuSimple score was high: 0.93 (left), 0.94 (right), while as can be observed there are many erroneous segments. In the right example, at the far range there are clearly large lateral offsets between the detections and the gt. However, since the tuSimple evaluation is in the image plane, these offsets become very small. In the right example, there are many small false segments on the far left and far right sides which the clustering manages to filter out.} 
    \label{fig:tuSimpleEval}
\end{figure}

\subsubsection{Segment-based evaluation}

To complete the description of the proposed segment-based evaluation described in Section~\ref{sec:experiments} we define the distance $seg\_dist(\textbf{p},\textbf{q})$ between two segments in the plane. The idea was to try to match lane segments that belong to the same lane, and therefore there is an emphasis on their orientation, and different treatment of the distance \textit{along} the segment direction and \textit{perpendicular} to it. 

Let $l_p,l_q$ be the corresponding infinite lines on which $p,q$ reside, and let ($(p^1,p^2),(q^1,q^2)\in\mathbb{R}^2$  be the segment endpoints. We project each endpoint to the opposite line (i.e. $p^1$ is projected to $l_q$ and denoted by $p^1_q$) to generate four projected points: $(p^1_q,p^2_q),(q^1_p,q^2_p)\in\mathbb{R}^2$. We start by eliminating some of the matches if the projected segment doesn't sufficiently overlap with the opposite segment: $\frac{|p^1_q p^2_q|}{|q^1q^2|}>0.5$ or $\frac{|q^1_p q^2_p|}{|p^1p^2|}>0.5$. Matches are eliminated by setting $seg\_dist(\textbf{p},\textbf{q})=\infty$. Finally, for the remaining matching pairs, distance is computed as the maximum distance between the end-points and their projected counterparts: $seg\_dist=\max(|p^1p^1_q|,|p^2p^2_q|,|q^1q^1_p|,|q^2q^2_p|)$.

\subsubsection{tuSimple evaluation}

As mentioned in Section~\ref{sec:experiments} for the tuSimple evaluation we need to cluster the segments. For this purpose we apply a heuristic clustering algorithm, operating on the tile representation output. Initially, in each row we apply a 1D non-maxima suppression with kernel size 20cm to suppress redundant detections of the same lane in neighboring tiles. 

The clustering progresses row by row from bottom to top. For each segment in a top row, it is connected to its 3 closest neighbors in the previous (bottom) row. Per connection, an affinity score is computed. The affinity measures the likelihood that the two segments belong to the same lane, using the following parameters. $\theta$ - the orientation difference between the segments, $d_{min}$ - the minimum euclidean distance between their endpoints,  $b_{curr}$ - confidence score for the current segment and $b_{prev}$ - confidence score of the connected neighbor from the previous row. If $\theta>45$ then the affinity $a=0$, otherwise, it is computed as: $a = b_{curr}\cdot b_{prev}\cdot \cos(\theta)\cdot\frac{8-d_{min}}{8}$. We then cluster the current segment with its highest scoring neighbor. For each cluster $b_{max}$ is set as the maximum score over segments in the cluster.

We next apply a filtering stage dropping clusters with less than four segments, and clusters with $b_{max}<10^{-2}$. Finally, we loop over all remaining cluster pairs and merge a pair if one cluster starts on the same row as the other one ends on horizontally-adjacent tiles. We continue untill no pairs can be merged. As can be seen in Figure~\ref{fig:tuSimpleEval}, due to the clustering, and the forgivingness of the tuSimple evaluation, especially at the farther ranges, the final metric is not very indicative of the quality of the raw tile-representation output.

\subsection{Additional  examples}
\begin{figure}[h]
    \centering
    \setlength\tabcolsep{2pt}
    \begin{tabular}{|c||c|c|c|}
    \hline
    
     & \textbf{Autoencoder} & \textbf{Image translation} & \textbf{Self-Supervision}\\
    \hline
    \hline
     \rotatebox[origin=c]{90}{\makecell{ \textbf{3DLanes} \hspace{-3cm}}} &
     \includegraphics[height=4cm]{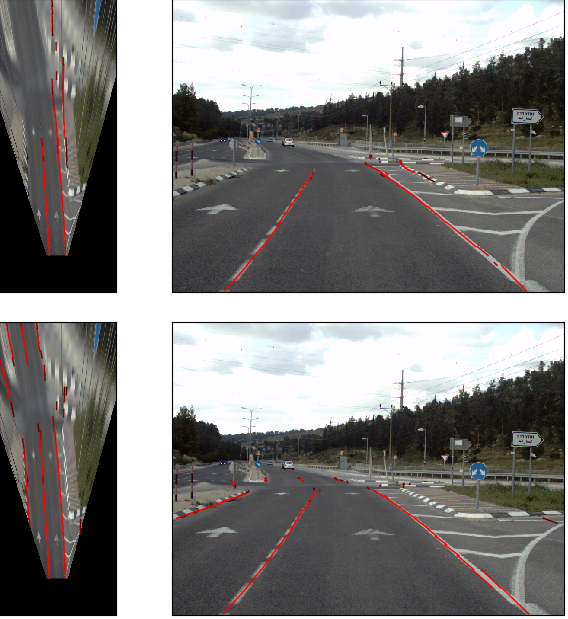} &  \includegraphics[height=4cm]{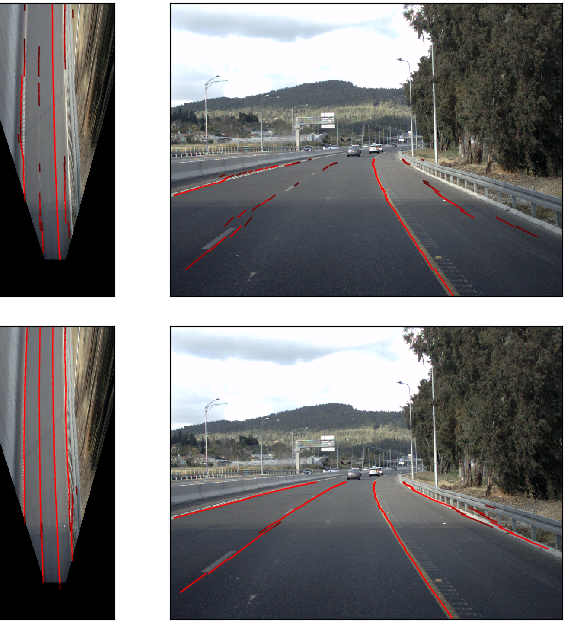} & 
    \includegraphics[height=4cm]{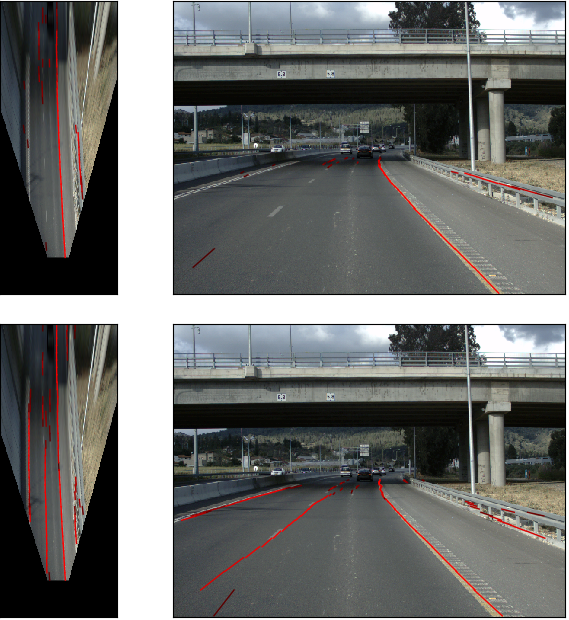}   \\
    \hline
    \hline
     \rotatebox[origin=c]{90}{\makecell{ \textbf{llamas} \hspace{-3cm}}} &
     \includegraphics[height=4cm]{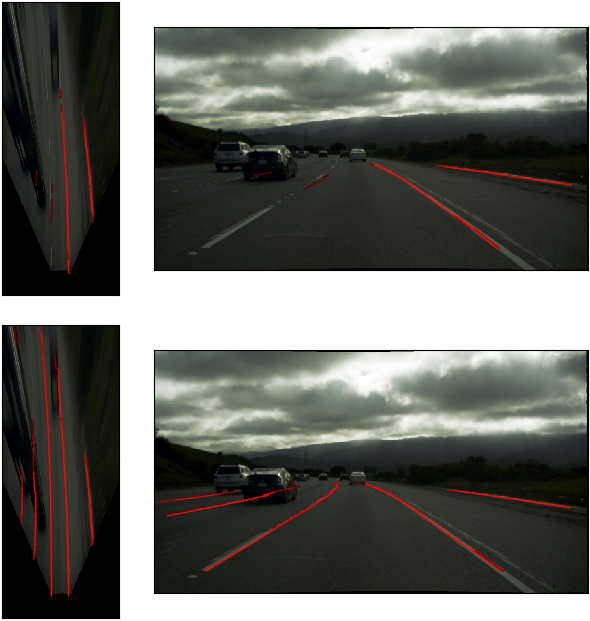} &  \includegraphics[height=4cm]{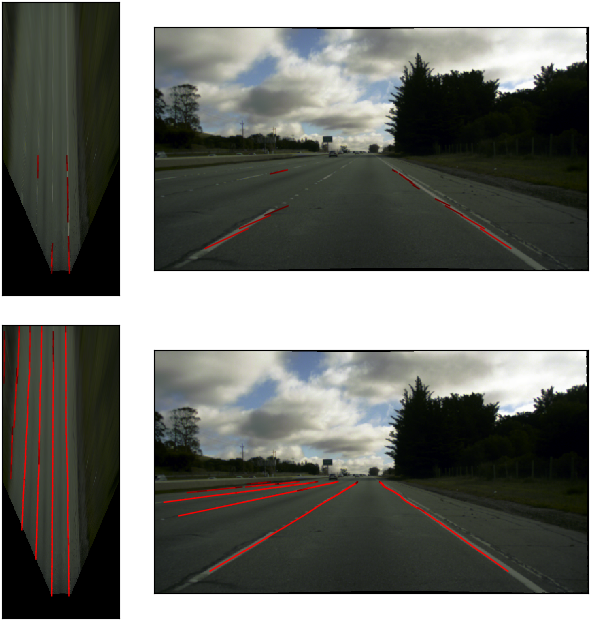} & 
    \includegraphics[height=4cm]{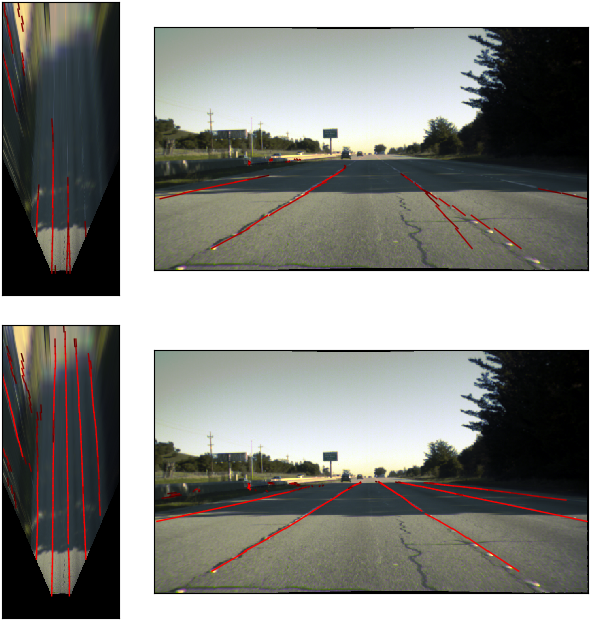}   \\
    \hline
    \hline
     \rotatebox[origin=c]{90}{\makecell{ \textbf{tuSimple} \hspace{-3cm}}} &
     \includegraphics[height=4cm]{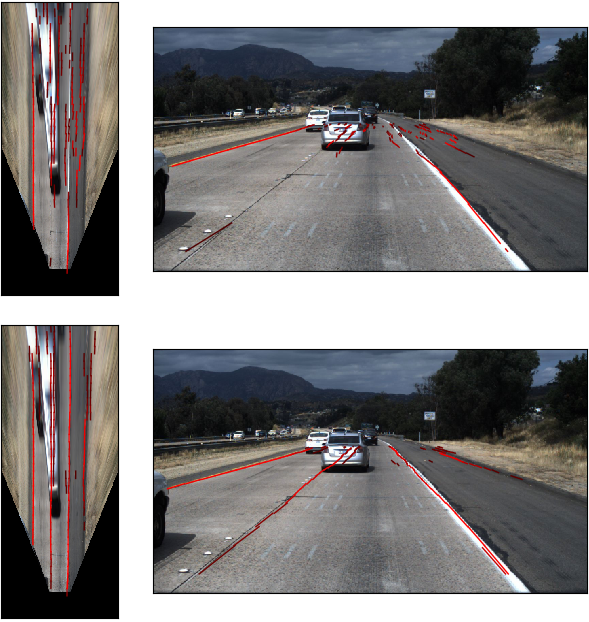} &  \includegraphics[height=4cm]{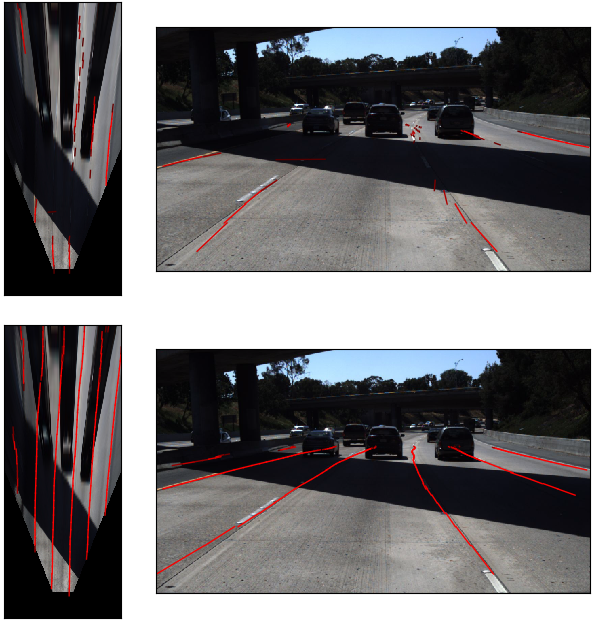} & 
    \includegraphics[height=4cm]{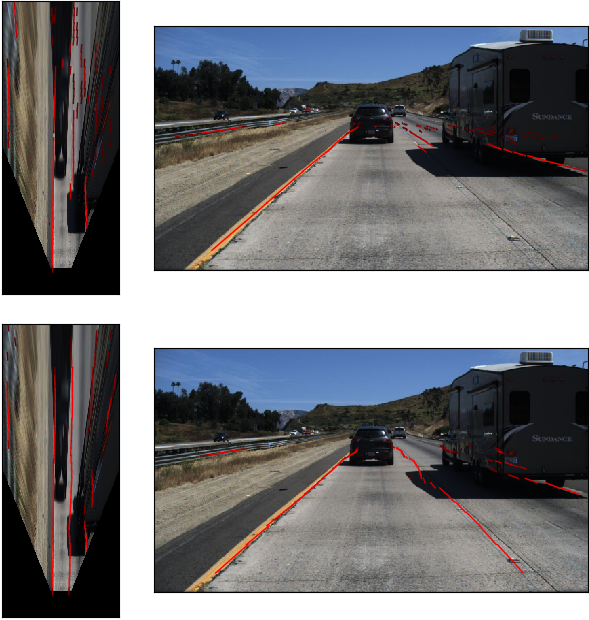}   \\
    \hline
    \end{tabular}         
    \caption{\textbf{Additional Results in the unsupervised domain adaptation (UDA) setting}. Sample results using the three leading methods (columns) on samples from the three different datasets we tested on (rows). \textbf{In each cell} we show the result before (top within cell) and after domain adaptation with the respective method (bottom). Results are shown in each cell in both top-view (right) and regular view (left).}
    \label{fig:additional_results}
\end{figure}

In Figure ~\ref{fig:additional_results} we show examples of additional results before and after domain adaptation in the UDA setting for the three leading methods, namely, image translation, self-supervision and our auto-encoder approach.

\subsection{Architectures}

Tables~\ref{tab:first}-\ref{tab:last} in this section specify the CNN architectures. Wherever “input\_name” is empty it is the output of the line above. "+" in the input means concatenation along channel dimension. All ReLUs are leaky relu with factor 0.1.

\begin{table}
\begin{tabular}{|l|llllll|}
\hline
type       & Cin & Cout & Kernel & Stride & input\_name & Output name \\
\hline
\hline
ConvBNRelu & 3   & 32   & 3      & 1      & bev image   &             \\
ConvBNRelu & 32  & 32   & 3      & 1      &             &             \\
maxpool    &     &      & 2      & 2      &             & embed\_2    \\
ConvBNRelu & 32  & 64   & 3      & 1      & embed\_2    &             \\
ConvBNRelu & 64  & 64   & 3      & 1      &             &             \\
maxpool    &     &      & 2      & 2      &             & embed\_4    \\
ConvBNRelu & 64  & 128  & 3      & 1      & embed\_4    &             \\
ConvBNRelu & 128 & 128  & 3      & 1      &             &             \\
ConvBNRelu & 128 & 128  & 3      & 1      &             &             \\
maxpool    &     &      & 2      & 2      &             & embed\_8    \\
ConvBNRelu & 128 & 128  & 3      & 1      & embed\_8    &             \\
ConvBNRelu & 128 & 128  & 3      & 1      &             &             \\
ConvBNRelu & 128 & 128  & 3      & 1      &             &             \\
maxpool    &     &      & 2      & 2      &             & embed\_16 \\ 
\hline
\end{tabular}
\caption{\textbf{Base architecture. }Image Embedding network, $\phi$.}\label{tab:first}
\end{table}

\begin{table}
\begin{tabular}{|l|llllll|}
\hline
type       & Cin & Cout                                      & Kernel & Stride & input\_name &  \\ \hline \hline
ConvBNRelu & 128 & 64                                        & 3      & 1      & embed\_16   &  \\
ConvBNRelu & 64  & 64                                        & 3      & 1      &             &  \\
ConvBNRelu & 64  & 64                                        & 3      & 1      &             &  \\
ConvBNRelu & 64  & 10  &  1 & 1 & & \\ \hline
\end{tabular}
\caption{\textbf{Base architecture.} Embedding to tile-representation network, $\psi$}
\end{table}

\begin{table}
\begin{tabular}{|l|llllll|}
\hline
type             & Cin & Cout & Kernel & Stride & input\_name                        &    output\_name            \\ \hline \hline
ConvBNRelu       & 128 & 256  & 3      & 1      & embed\_16                          &                    \\
ConvBNRelu256    & 256 & 256  & 3      & 1      &                                    &                    \\
ConvBNRelu       & 256 & 256  & 3      & 1      &                                    &                    \\
maxpool          &     &      & 2      & 2      &                                    & embed\_32          \\
ConvBNRelu       & 128 & 64   & 3      & 1      & embed\_16                          & embed\_16\_reduced \\
ConvBNRelu       & 128 & 64   & 3      & 1      & embed\_8                           & embed\_8\_reduced  \\
ConvBNRelu       & 64  & 32   & 3      & 1      & embed\_4                           & embed\_4\_reduced  \\
Nearest Upsample &     &      & 2      &        & embed\_32                          &                    \\
ConvBNRelu       & 256 & 64   & 3      & 1      &                                    &                    \\
ConvBNRelu       & 64  & 64   & 3      & 1      &                                    &                    \\
ConvBNRelu       & 64  & 64   & 3      & 1      &                                    & embed\_16\_up      \\
Nearest Upsample &     &      & 2      &        & embed\_16\_reduced +  &                    \\
 &     &      &       &        &  embed\_16\_up &                    \\
ConvBNRelu       & 128 & 64   & 3      & 1      &                                    &                    \\
ConvBNRelu       & 64  & 64   & 3      & 1      &                                    &                    \\
ConvBNRelu       & 64  & 64   & 3      & 1      &                                    & embed\_8\_up       \\
Nearest Upsample &     &      & 2      &        & embed\_8\_reduced +    &            \\      &     &      &       &        &  &embed\_8\_up                       \\
ConvBNRelu       & 128 & 32   & 3      & 1      &                                    &                    \\
ConvBNRelu       & 32  & 32   & 3      & 1      &                                    &                    \\
ConvBNRelu       & 32  & 32   & 3      & 1      &                                    & embed\_4\_up       \\
ConvBNRelu       & 64  & 64   & 3      & 1      & embed\_4\_reduced  +  &                    \\
       &   &    &       &       & embed\_4\_up   &                    \\
ConvBNRelu       & 64  & 64   & 3      & 1      &                                    &                    \\
ConvBNRelu       & 64  & 64   & 3      & 1      &                                    &                    \\
ConvBNRelu       & 64  & 1    & 3      & 1      &                                    & l - “lanes image” \\
\hline
\end{tabular}
\caption{\textbf{Autoencoder.} embedding to skeleton network, $\delta$}
\end{table}

\begin{table}
\begin{tabular}{|l|llllll|}
\hline
type                             & Cin                                   & Cout & Kernel & Stride & input\_name                                     &    output\_name         \\
\hline
\hline
convSpectralNormRelu             & 10  & 64   & 4      & 2      & “lanes image”  &             \\
convSpectralNormInstanceNormRelu & 64                                    & 128  & 4      & 2      &                                                 &             \\
convSpectralNormInstanceNormRelu & 128                                   & 256  & 4      & 2      &                                                 &             \\
convSpectralNormInstanceNormRelu & 256                                   & 512  & 4      & 2      &                                                 &             \\
Conv                             & 512                                   & 128  & 4      & 1      &                                                 & d\_internal \\
Minibatch discrimination layer   & 128                                   & 128  &        &        & d\_internal                                     & md          \\
Conv                             & 256                                   & 1    & 1      & 1      & d\_internal +                             &   \\         
                             &                                    &     &       &      &  md                                &   \\
                             \hline
                             
\end{tabular}
\caption{\textbf{Autoencoder.} Discriminator network, $\gamma$. For  \textbf{Embedding GAN} the discriinator is identical, and the input is $embed\_16$ with channel size is 128. Spectral Norm is described in~\cite{miyato2018spectral} and the Minibatch discrimination layer in~\cite{SalimansGZCRC16}}.
\end{table}

\begin{table}
\begin{tabular}{|l|lllll|}
\hline
type             & Cin & Cout & Kernel & Stride & input\_name  \\
\hline
\hline
ConvBNRelu       & 1   & 16   & 3      & 1      & “lane image” \\
ConvBNRelu       & 16  & 32   & 3      & 1      &              \\
ConvBNRelu       & 32  & 64   & 3      & 1      &              \\
Nearest Upsample &     &      & 2      &        &              \\
ConvBNRelu       & 64  & 64   & 3      & 1      &              \\
ConvBNRelu       & 64  & 64   & 3      & 1      &              \\
ConvBNRelu       & 64  & 64   & 3      & 1      &              \\
Nearest Upsample &     &      & 2      &        &              \\
ConvBNRelu       & 64  & 64   & 3      & 1      &              \\
ConvBNRelu       & 64  & 64   & 3      & 1      &              \\
ConvBNRelu       & 64  & 64   & 3      & 1      &              \\
ConvBNRelu       & 64  & 1    & 3      & 1      &             \\
\hline
\end{tabular}
\caption{\textbf{Autoencoder.} Skeleton to gradient image network, $\lambda$.}
\end{table}

\begin{table}
\begin{tabular}{|l|lllll|}
\hline
type             & Cin & Cout & Kernel & Stride & input\_name  \\
\hline\hline
ConvBNRelu & 128 & 64 & (5, 3) & 1 & embed\_16 \\
maxpool    &     &    & 2      & 2 &           \\
ConvBNRelu & 64  & 64 & (5, 3) & 1 &           \\
maxpool    &     &    & 2      & 2 &           \\
conv       & 64  & 3  & 1      & 1 &          \\
\hline
\end{tabular}
\caption{\textbf{Self-supervision}. Auxiliary task classifier, $\rho$.}
\label{tab:last}
\end{table}

\end{document}